\documentclass[twoside]{article}

\usepackage[accepted]{aistats2020}

\usepackage{amsmath}
\usepackage{mathtools}
\usepackage{algorithmic}
\usepackage{algorithm}
\usepackage{placeins}
\usepackage{bbold}

\DeclareMathOperator*{\argmax}{argmax}

\DeclareMathOperator{\FC}{FC}
\DeclareMathOperator{\Conv}{Conv}
\DeclareMathOperator{\avg}{\hat{\mathbb{E}}}
\DeclarePairedDelimiter\floor{\lfloor}{\rfloor}

\newcommand{\gr}{\mathcal{G}}
\newcommand{\p}[1]{#1^{\prime}}
\newcommand{\old}[1]{#1^{-}}

\newcommand{\states}{\mathcal{S}}
\newcommand{\actions}{\mathcal{A}}
\newcommand{\norm}[1]{\left\lVert#1\right\rVert}
\newcommand{\abs}[1]{\left\lvert#1\right\rvert}

\usepackage[short]{optidef}

\usepackage[utf8]{inputenc} %
\usepackage[T1]{fontenc}    %
\usepackage{hyperref}       %
\usepackage{amsthm}
\usepackage{url}            %
\usepackage{booktabs}       %
\usepackage{amsfonts}       %
\usepackage{nicefrac}       %
\usepackage{microtype}      %
\usepackage{mathtools}
\usepackage{authblk}
\usepackage{lmodern}
\usepackage{framed}
\usepackage{natbib}

\mathtoolsset{showonlyrefs=true}

\newtheorem{theorem}{Theorem}
\newtheorem{proposition}{Proposition}
\newtheorem{corollary}{Corollary}
\newtheorem{lemma}{Lemma}
\newtheorem{assumption}{Assumption}

\begin{document}
\twocolumn[

\aistatstitle{Momentum in Reinforcement Learning}

\aistatsauthor{Nino Vieillard$^{1,2}$\And Bruno Scherrer$^2$ \And  Olivier Pietquin$^1$ \And Matthieu Geist$^1$}
\runningauthor{Nino Vieillard, Bruno Scherrer, Olivier Pietquin, Matthieu Geist}
\aistatsaddress{ $^1$Google Research, Brain Team \\ $^2$Université de Lorraine, CNRS, Inria, IECL, F-54000 Nancy, France} ]

\begin{abstract}
    We adapt the optimization's concept of momentum to reinforcement learning. Seeing the state-action value functions as an analog to the gradients in optimization, we interpret momentum as an average of consecutive $q$-functions. We derive Momentum Value Iteration (MoVI), a variation of Value iteration that incorporates this momentum idea. Our analysis shows that this allows MoVI to average errors over successive iterations. We show that the proposed approach can be readily extended to deep learning. Specifically,we propose a simple improvement on DQN based on MoVI, and experiment it on Atari games.
\end{abstract}

\section{Introduction}
Reinforcement Learning (RL) is largely based on Approximate Dynamic Programming (ADP), that provides algorithms to solve Markov Decision Processes (MDP, \citet{puterman1994markov}) under approximation.  In the exact case, where there is no approximation, classic algorithms such as Value Iteration (VI) or Policy Iteration (PI) are guaranteed to converge to the optimal solution, that is find an optimal policy that dominates every policy in terms of value. These algorithms rely on solving fixed-point problems: in VI,  one tries to reach the fixed point of the Bellman optimality operator by an iterative method. We focus on VI for the rest of the paper, but the principle we propose can be extended beyond this. Approximate Value Iteration (AVI) is a VI scheme with approximation errors. It is well known \citep{bertsekas1996neuro} that if the errors do not vanish, AVI does not converge. To get some intuition, consider a sequence of policies being greedy according to the optimal $q$-function, with an additional state-action dependant noise. The resulting sequence of policies will be unstable and suboptimal, even with centered and bounded noise.  Dealing with errors is however crucial to RL, as we hope to tackle problems with large states spaces that require function approximation. Indeed, many recent RL successes are algorithms that instantiate ADP schemes with neural networks for function approximation. Deep Q-Networks (DQN, \citet{mnih2015human}) for example, can be seen as an extension of AVI with neural networks. 

In optimization, a common strategy to stabilize the descent direction, known as momentum, is to average the successive gradients instead of considering the last one. In reinforcement learning, the state-action value function can be seen informally as a kind of gradient, as it gives an improvement direction for the policy. Hence, we propose to bring the concept of momentum to reinforcement learning by basically averaging $q$-values in a DP scheme.

We introduce Momentum Value Iteration (MoVI) in Section~\ref{sec:movi}. It is Value Iteration, up to the fact that the policy, instead of being greedy with respect to the last state-action value function, is greedy with respect to an average of the past value functions. We analyze the propagation of errors of this scheme. In AVI, the performance bound will depend on a weighted sum of the norms of the errors at each iteration. For MoVI, we show that this depends on the norms of the cumulative errors of previous iteration. This means that it allows for a compensation of errors along different iterations, and even convergence in the case of zero-mean and bounded noises, under some assumption. This compensation property is shared by a few algorithms that will be discussed in Section \ref{sec:related}. We also show that MoVI can be successfully combined with powerful function approximation by proposing Momentum-DQN in Section~\ref{sec:modqn}, an extension of MoVI with neural networks based on DQN. It provides a strong performance improvement over DQN on the standard Arcade Learning Environment (ALE) benchmark \citep{bellemare2013arcade}. All stated results are proven in the appendix.

\section{Background}
\label{sec:background}

\paragraph{Markov Decision Processes.} We consider the RL setting  where an agent interacts with an environment modeled as an infinite discounted horizon MDP. An MDP is a quintuple $\{\states, \actions, P, r, \gamma\}$, where $\states$ is a finite\footnote{This is for ease and clarity of exposition, the proposed algorithm and analysis can be extended to continuous state spaces.} state space, $\actions$ a finite action space,  $P \in\Delta_\states^{\states\times\actions}$ is a Markovian transition kernel (writing $\Delta_X$ the simplex over the set $X$), $r \in  [-r_{\max} , r_{\max}]^{\states\times\actions}$ a reward function and $\gamma \in (0,1)$ the discount factor. A policy $\pi$ maps the state space to distributions over actions $\pi(\cdot | s)$. We define the $q$-value $q_\pi$ of a policy $\pi$ as, for each $s \in \states$ and $a \in \actions$,
\begin{equation}
    q_\pi(s, a) = \mathbb{E}_{\pi}\left[\sum_{t=0}^{\infty} \gamma^t r(s_t, a_t) \,\middle\vert\, s_0=s, a_0=a \right],
\end{equation}
where $\mathbb{E}_{\pi}$ denotes the expected value over all trajectories $(s_1, a_1, s_2, a_2, \hdots)$ produced by $\pi$. The value is bounded by $q_{\max} = r_{\max}/(1-\gamma)$. Let us define the transition kernel operator associated to $\pi$ as, for each $q \in \mathbb{R}^{\states\times\actions}$ and for each $(s,a) \in \states\times\actions$, as $[P_\pi q](s,a) = \mathbb{E}_{\p{s}\sim P(\cdot|s,a), \p{a}\sim \pi(\cdot|\p{s})}[q(\p{s}, \p{a})]$. The $q$-function of a policy is the fixed point of its Bellman evaluation operator, defined for each $q \in \mathbb{R}^{\states\times\actions}$ as $T_\pi q = r + \gamma P_\pi q$. An optimal policy $\pi_*$ is such that for any other policy $\pi$, we have that, for each $(s,a)\in\states\times\actions$, $q_{\pi_*}(s,a) \geq q_\pi(s,a)$. The Bellman optimality operator is defined as $T_* q = \max_\pi T_\pi q$, and we have that $q_*$ is the unique fixed point of $T_*$. A policy is said to be greedy with respect to $q \in \mathbb{R}^{\states\times\actions}$ if  $T_*q = T_\pi q$. We denote the set of these policies $\gr(q)$. Note that such a policy can be computed without accessing to the model (the transition kernel).

Finally, for $\mu\in\Delta_{\states\times\actions}$ we write $d_{\pi,\mu} = (1 -\gamma)\mu(I-\gamma P_{\pi})^{-1}$ the discounted cumulative occupancy measure induced by $\pi$ when starting from the distribution $\mu$ (distributions being written as row vectors). We define the $\mu$-weighted $\ell_p$-norm as, for each $q\in \mathbb{R}^{\states\times\actions}$, $\norm{q}_{p,\mu} = \left(\mathbb{E}_{(s,a)\sim \mu}\left[\abs{q(s,a)}^p\right]\right)^{\frac{1}{p}}$.

\paragraph{Approximate Value Iteration.} Approximate Dynamic Programming provides algorithms to solve an MDP under some errors. One classic algorithm is Approximate Value Iteration. It looks directly for the fixed point of $T_*$ with an iterative process
\begin{equation}
\label{eq:avi}
    \begin{cases}
    \pi_{k+1} \in \gr(q_k) \\
    q_{k+1} = T_{\pi_{k+1}} q_k + \epsilon_{k+1}.
    \end{cases}
    \quad \text{(AVI)}
\end{equation}
Notice that here, $T_{\pi_{k+1}} q_k = T_* q_k$. In this scheme, we call the first line the greedy step, and the second line the partial evaluation step. AVI satisfies the following bound for the quality of the policy $\pi_k$
\begin{equation}
\label{eq:vi-bound}
    \lVert q_* - q_{\pi_k} \rVert_\infty \leq 2\gamma^{k}q_{\max} +  \frac{2\gamma \max_{j<k} \norm{\epsilon_j}_\infty}{(1 - \gamma)^2}.
\end{equation}
This explains why AVI is not resistant to errors: $\max_{j<k} \norm{\epsilon_j}_\infty$ can be high even if each $\epsilon_k$ is zero-mean.

\section{Momentum Value Iteration}
\label{sec:motivation}
In the context of optimization, momentum aims at stabilizing gradient ascent (or descent) methods. Consider we want to maximize a concave function $f$ whose gradient is not known analytically, and we use a classic (stochastic) gradient ascent algorithm. This algorithm iterates from a value $x_0$ by computing an approximation  $g_k$ of $\nabla f(x_k)$, and updating $x_{k+1} = x_k + \eta g_k$. One can then use momentum \citep{qian1999momentum} to stabilize the process through a smoothing function $h_k = \rho h_k + g_k$, with $\rho \in \mathbb{R}$, and an update $x_{k+1} = x_k + \eta h_k$. This can stabilize the ascent as the gradient may vary greatly from step to step. 

In the context of ADP, the $q$-function intuitively gives the direction that guides the policy, in the same way that the gradient is the improvement direction of a variable. In particular, we can rewrite the greedy step (in AVI) as $\pi_k(\cdot|s) \in  \argmax_{\pi(\cdot|s) \in \Delta_\actions} \langle q_k(s,\cdot), \pi(\cdot|s) \rangle$, thus seeing this step as finding the policy being state-wise the most colinear with $q_k$. This is also reminiscent of the direction finding subproblem of~\citet{frank1956algorithm}. Consequently, the greedy step can be seen as an analog of the update in gradient ascent (the policy $\pi$ is analog to the variable $x$), the differences being  \textit{(i)} that $q_k$ in AVI is not a gradient, but the result of an iterative process, $q_k = T_{\pi_{k}} q_{k-1}$, and \textit{(ii)} that the policy is not updated, but replaced. 

This analogy is thus quite limited ($q_k$ is not really a gradient, there is no optimized function, the policy is replaced rather than updated). However, it is sufficient to adapt the momentum idea to AVI, by replacing the $q$-function in the improvement step by a smoothing of the $q$-functions, $h_k = \rho h_{k-1} + q_k$. We can then notice that $\gr(h_k) = \gr(\frac{h_k}{1+\rho})$, allowing us to compute a moving average instead of a smoothing, $h_k = \beta_k h_{k-1} + (1 - \beta_k)q_k$, which leads to the following ADP scheme, initialized with $h_0=q_0$,
\begin{equation}
\label{eq:movi}
    \begin{cases}
    \pi_{k+1} = \gr(h_k) \\
    q_{k+1} = T_{\pi_{k+1}} q_k + \epsilon_{k+1}  \\
    h_{k+1} = \beta_{k+1} h_{k} + (1 - \beta_{k+1})q_{k+1} .
    \end{cases}
    \quad \text{(MoVI)}
\end{equation}
We call this scheme Momentum Value Iteration (MoVI), we analyze it in the following section.

\section{Analysis}
\label{sec:movi}
For the analysis, we consider a specific case of the scheme in Equation~\eqref{eq:movi}, with an empirical mean rather than an iteration-dependant moving average. This amounts to define $\beta_k=\frac{k}{k+1}$ in Eq.~\eqref{eq:movi}. We study the propagation of errors of MoVI, to see how it is impacted by the introduction of momentum, compared to a classic AVI scheme (see Eq.~\eqref{eq:vi-bound}).

\subsection{Error propagation analysis}
First, let us define some useful notations. We denote $P_{j:i} = P_{\pi_j}P_{\pi_{j-1}} \hdots P_{\pi_i}$ if $1 \leq i \leq j$, $P_{j:i} = I$ otherwise, where $\pi_j$ is the policy computed by MoVI at iteration $j$. We then define the negative cumulative error $E_{k} = - \sum_{j=1}^{k}\epsilon_j$, and the weighted negative cumulative error $\p{E}_{k,j} = -\sum_{i = 1}^{k-j} P_{i+j:i+1}(I - \gamma P_{{\pi}_i}) \epsilon_i$. 

To study the efficiency of the algorithm, the natural quantity to bound is the loss $q_* - q_{\pi_k}\geq 0$, the difference between the value of the optimal policy and the (true) value of the policy computed by MoVI.

\begin{theorem}
\label{th:bound}
After $k+1$ iterations of MoVI, we have
\begin{multline}
    q_* - q_{\pi_{k+1}}  \leq \frac{1}{k+1} \Bigg[ (I - \gamma P_{\pi_*})^{-1} (E_{k+1} + q_{k+1} - q_0) \\
                         - (I - \gamma P_{\pi_{k+1}})^{-1} \bigg(\sum_{j=0}^{k-1} \gamma^j\p{E}_{k,j}
                            +\sum_{j=0}^k \gamma^j P_{j:1}(T_{\pi_1} q_0 - q_0)\bigg) \Bigg].
\end{multline}
\end{theorem}

To understand and then discuss this result, we provide a bound of a $\mu$-weighted $\ell_1$-norm of the loss: the norm is what one would want to control in a practical setting. Notice that we could similarly derive a bound for the $\mu$-weighted $\ell_p$-norm.

\begin{corollary}
\label{th:bound-1mu}
Let $\mu$ be the distribution of interest, and $\nu$ the sampling distribution. We introduce the following concentrability coefficient (the fraction being componentwise)
\begin{equation}
 C = \max_{\pi} \norm{\frac{d_{\pi, \mu}}{\nu}}_\infty
\end{equation}
Suppose that  we initialize $h_0 = q_0 = 0$. At iteration $k+1$ of MoVI, we have
\begin{multline}
\Vert q_* - q_{\pi_{k+1}} \Vert_{1, \mu} \leq \frac{C}{(k+1)(1-\gamma)} \Bigg( \Vert E_{k+1} \Vert_{1,\nu} +\\
											  \sum_{j=0}^{k-1} \gamma^j \Vert \p{E}_{k,j} \Vert_{1,\nu} + 2 q_{\max} \Bigg)	.										  
\end{multline}
\end{corollary}

Theorem \ref{th:bound} shows that $q_* - q_{\pi_k}$ depends on two error terms, $E_k$ and a $\gamma$-discounted sum of $\p{E}_{k,j}$. The first term corresponds to a sum of errors, that can then compensate, which is not the case in AVI (see Equation~\eqref{eq:vi-bound}). The normalization by $\frac{1}{k+1}$ reduces the variance of this term, and that can lead to convergence under some assumptions (see Section~\ref{sec:sample}). However, the second term is more cumbersome. The terms $\p{E}_{k,j}$ depend on sums of error weighted by composed kernels $P_{i:j}$. Would these kernels be arbitrary, this could lead to further variance reduction. However, the corresponding average is done over the state-action space in addition to over iterations, and the kernels are dependent of the error they weight, this dependency being hard to quantify. We further discuss this next.

\begin{figure}
    \centering
    \includegraphics[width=\linewidth]{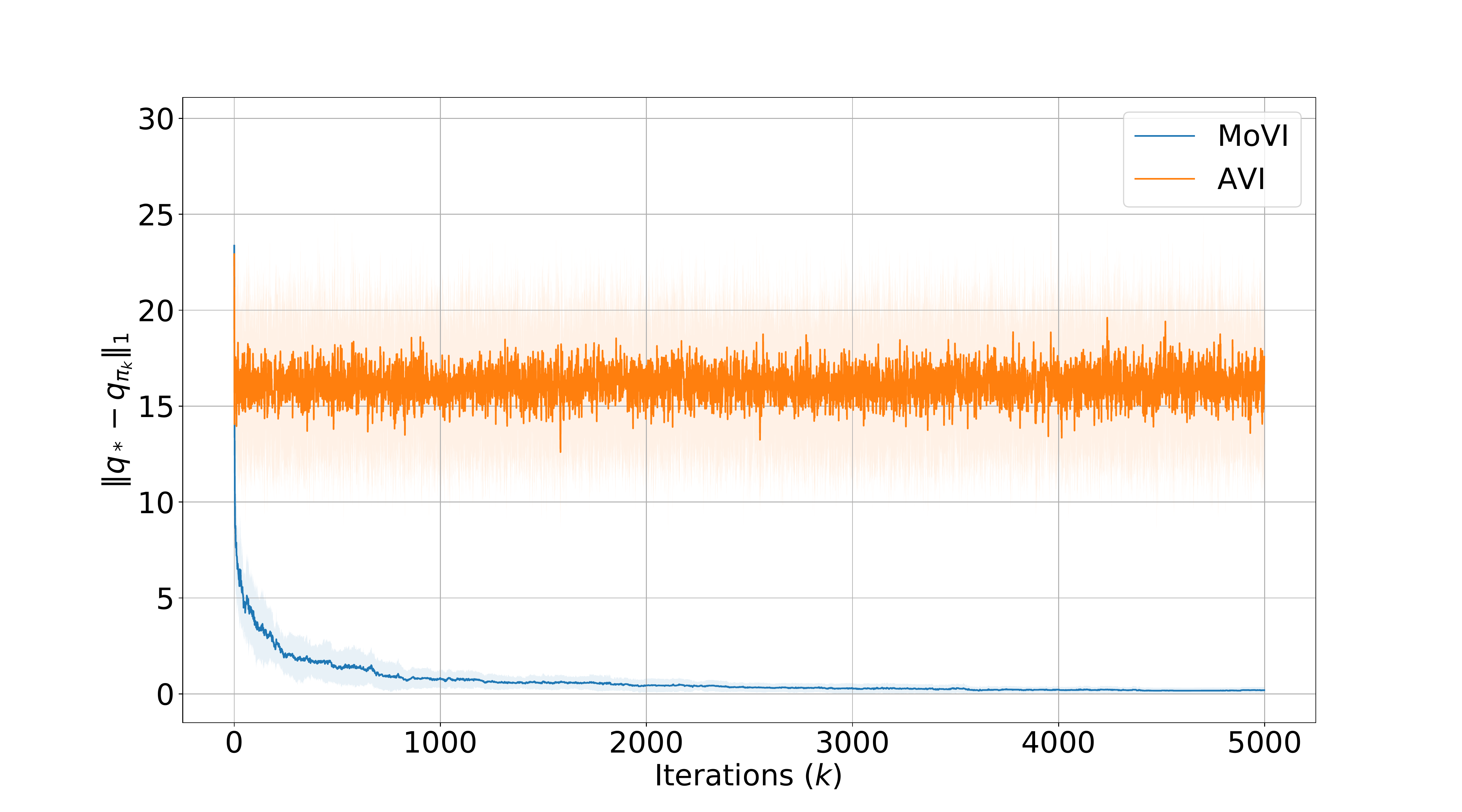}
    \caption{Illustration of the convergence of MoVI. We represent the empirical mean and standard deviation of the error over $100$ MDPs.} 
    \label{fig:movi-bound}
\end{figure}

Still, the algorithm can converge in practice, and we illustrate its behaviour on a simple case. We give ourselves a tabular representation of a randomly generated  MDP, with access to a generative model. The approximation comes from the fact that the Bellman operator is sampled at each iteration (instead of being evaluated exactly); we compare it to AVI in the same scenario. We report the average error between $q_{\pi_k}$ and $q_*$ in Figure~\ref{fig:movi-bound}. This experiments illustrate how AVI oscillates with high error, while MoVI converges to $q_*$.

We note that our proof technique should hold with a constant $\beta$ too (moving average instead of average). In this case, instead of having an average error ($k^{-1}E_k$), we would have a moving average of the (weighted) errors. This would not vanish asymptotically, even with zero-mean bounded noises $\epsilon_k$, but this would still reduce the variance, and improve upon the AVI bound.

\subsection{About the sample complexity}
\label{sec:sample}
To better understand MoVI, we analyze its sample complexity in a simple case, Sampled-MoVI. In this setting, we have access to a generative model of the MDP and we give ourselves a tabular representation of the MDP. At each iteration of Sampled-MoVI, for each $(s,a) \in \states\times\actions$, we sample a state $\p{s} \sim P(\cdot|s,a)$ and perform the update from Equation~\eqref{eq:movi} with only this state. We denote $\hat{T}_{\pi_k}$ the resulting sampled Bellman operator, $\hat{T}_{\pi_k}q(s,a) = r(s,a) + \gamma q(\p{s},\pi_k(s))$. The error at iteration $k$ is then, for each $(s,a)$, $\epsilon_k(s,a) = \hat{T}_{\pi_k}q_{k-1}(s,a) - T_{\pi_k}q_{k-1}(s,a)$. It is thus zero-mean and centered. We provide a detailed pseudo-code in the Appendix.

We are interested in controlling the distance of our policy to the optimal policy, precisely in the norm $\Vert q_* - q_{\pi_k}\Vert_\infty$ at iteration $k$ of Sampled-MoVI. We have, as a direct consequence of Thm.~\ref{th:bound}, that
\begin{multline}
\label{eq:supremum}
    \Vert q_* - q_{\pi_{k+1}} \Vert_{\infty} \leq \frac{1}{(k+1)(1-\gamma)} \Bigg( \Vert E_{k+1} \Vert_{\infty} +\\
											  \sum_{j=0}^{k-1} \gamma^j \Vert \p{E}_{k,j} \Vert_{\infty} + 2q_{\max} \Bigg).					
\end{multline}
Informally, using an Hoeffding argument, we have $k^{-1}\|E_k\|_\infty = \mathcal{O}(k^{-\frac{1}{2}})$. However, bounding a term $\max_{j\leq k} \Vert \p{E}_{k,j} \Vert_\infty$ is more involved. This could typically be done using the Maximal Azuma-Hoeffding inequality. Yet, this requires the errors to be centered and bounded. In our case, the sequence of estimation errors $\{\epsilon_1(s,a), \hdots \epsilon_k(s,a)\}$ is a martingale difference sequence with respect to the natural filtration $\mathcal{F}_{k-1}$ (generated by the sequence of states sampled from the generative model), that is $\mathbb{E}[\epsilon_k(s,a) | \mathcal{F}_{k-1}] = 0$. This is sufficient for controlling the term $E_k$, but the terms $\p{E}_{k,j}$ are more difficult. Indeed, there, the errors are multiplied by a series of transition kernel matrices. For an arbitrary kernel $P$, independent of $\epsilon_k$, we would have $\mathbb{E}[P\epsilon_k(s,a) | \mathcal{F}_{k-1}] = P\mathbb{E}[\epsilon_k(s,a) | \mathcal{F}_{k-1}] = 0$. Unfortunately $P_{\pi_{k+1}}$ depends on $\pi_{k+1}$, which is greedy with respect to $h_k$, which is computed using $q_k$ and so depends on $\epsilon_k$. Thus, the independence cannot be assessed. To control the error in Sampled-MoVI, we consequently make the following assumption.
\begin{assumption} 
\label{co:filtration}
$\forall i,j \geq 1$, $\mathbb{E}\left[P_{j+i:j+1}\epsilon_j \vert \mathcal{F}_{j-1}\right] = 0$.
\end{assumption}
This assumption may seem very strong, as the dependency is hard to quantify. However, we have that $\pi_{k+1} \in \gr(h_k) =  \gr(\frac{k}{k+1} h_{k-1} + \frac{1}{k+1} T_{\pi_{k}} q_{k-1} + \frac{1}{k+1} \epsilon_k$). Thus, the influence of $\epsilon_k$ on $\pi_k$ diminishes with time.
Indeed, assuming that $\mathbb{E}\left[P_{j+i:j+1}\epsilon_j \vert \mathcal{F}_{j-1}\right] = o(\frac{1}{\sqrt{j}})$ should be enough to ensure convergence, but at a lower speed. We study numerically this assumption in Section~\ref{sec:experiments}.

\begin{proposition}
\label{th:sc}
Suppose Asm.~\ref{co:filtration} holds. After $k$ iterations of Sampled MoVI, with probability at least $1-\delta$
\begin{equation}
    \lVert q_* - q_{\pi_{k}} \rVert_\infty \leq \frac{2r_{\max}}{(1-\gamma)^{2}} \Bigg[\frac{1}{k} + \frac{3}{(1-\gamma)}
                                                                                    \sqrt{ \frac{2 \ln{\frac{4\abs{\states}\abs{\actions}}
                                                                                                             {\delta}}
                                                                                                }{k}}
                                                                  \Bigg].
\end{equation}
\end{proposition}
This result only holds under the strong Asm.~\ref{co:filtration}. Under this setting (tabular representation, generative model), there exist algorithms with faster convergence~\citep{wainwright2019variance}. However, they are not easily extandable beyond this setting, contrary to MoVI that can be easily turned into a practical large scale deep RL algorithm.

\section{Momentum DQN}
\label{sec:modqn}
We now propose an extension of MoVI to Momentum-DQN, introducing stochastic approximation and using deep neural networks for function approximation. We base ourselves on Deep Q-Networks (DQN , \citet{mnih2015human}), using the same algorithmic structure. We propose an off-policy algorithm, using a replay buffer as in DQN: we can apply the Bellman evaluation operator to the estimated $q$-function in an off-policy manner.

We parametrize the $q$-function by an \textit{online} network $Q_\theta$ of weights $\theta$, and we keep a copy of these weights in a \textit{target} network $\old Q$ of weights $\old\theta$. We additionally define the averaging network $H_\phi$ of weights $\phi$, and their target counterparts $\old H$ and $\old\phi$. Momentum-DQN interacts in an online way with an environment collecting transitions $\{s,a,r, \p s\} \in \states \times \actions \times \mathbb{R} \times \states$, that are stored in a FIFO replay buffer $\mathcal{B}$. In DQN, the algorithm performs gradient descent to approximate the partial evaluation step by regressing an approximation of $T_* \old{Q}$, and periodically copies the weights of the online networks to the target networks. The loss minimized at each step is almost the same as in DQN, replacing an approximation of $T_*Q_k$ by an approximation of the evaluation operator of the greedy policy with respect to the averaging network, $T_{\gr(H_\phi)}\old{Q}$. We define a regression target for $Q_\theta$ as
\begin{equation}
    \hat{Q}(r,\p{s}) = r + \gamma \old{Q}(\p{s}, \argmax \old{H}(\p{s}, \cdot)), 
\end{equation}
and a regression loss 
\begin{equation}
\label{eq:qtarget}
     \mathcal{L}_q(\theta) = \avg_{\mathcal{B}} \left[ \left(\hat{Q}(r,\p{s}) - Q_\theta(s,a)\right)^2  \right],
\end{equation}
with $\avg$ the empirical loss over a finite set. Then, we define a regression loss for the averaging network as an approximation of Equation \eqref{eq:movi}. We use the general scheme from Equation~\eqref{eq:movi} with a possibly variable mixture rate $\beta_k$. The regression target $\hat H$ for the averaging network is computed as 
\begin{equation}
    \hat{H}(s,a,r, \p{s}) = \beta_k \old{H}(s,a) + (1 - \beta_k)\hat{Q}(r,\p{s}),
\end{equation}
which leads to a regression loss
\begin{equation}
\label{eq:htarget}
      \mathcal{L}_H(\phi) = \avg_{\mathcal{B}} \left[ \left(\hat{H}(s,a,r, \p{s}) - H_\phi(s,a)\right)^2  \right].
\end{equation}
Momentum-DQN interacts with the environment with the policy $\gr_{e_k}(H)$  that is $e_k$-greedy with respect to $H$, the averaging network ($e_k$ depends on $k$ because  we use a classic decreasing schedule for the exploration). During training, it minimizes losses $\mathcal{L}_q$ and $\mathcal{L}_h$ with stochastic gradient descent (or a variant), and update the target weights with the online weights every $C$ gradient steps. A detailed pseudo-code is given in Algorithm \ref{algo:mom-dqn}, and we evaluate this algorithm %
in Section \ref{sec:atari}.

\paragraph{On the mixture rate.}
 We  aim at considering a rate close to the one of MoVI, $\beta_k = \frac{k}{k+1}$. Due to stochastic approximation, an iteration of Momentum-DQN does not match one iteration of MoVI, rather we should wait for several target updates before considering we have performed such an iteration. Consequently, we consider a rate such that $\beta_k = \frac{\floor{k/\kappa}}{\floor{k/\kappa}+1} $, with $\kappa$ a rate update period (an hyperparameter), that is the number of environment steps between each change of $\beta$.

\begin{algorithm}[tbh]
\caption{Momentum-DQN}
\begin{algorithmic}%
\label{algo:mom-dqn}
\REQUIRE $K\in \mathbb{N^*}$ the number of steps, $C\in \mathbb{N^*}$ the update period, $F \in \mathbb{N^*}$ the interaction period, $\kappa \in \mathbb{N^*}$ the rate update period.
\STATE Initialize $\theta$, $\phi$ at random
\STATE $\mathcal{B} = \{\}$
\STATE $\old\theta = \theta, \old\phi = \phi$
\FOR{$k = 1$ \TO $K$}
    \STATE Collect a transition $t = (s, a, r, \p{s})$ from $\gr_{e_k}(H_\phi)$
    \STATE $\mathcal{B} \leftarrow \mathcal{B} \cup \{t\}$
    \IF{$k \mod F == 0$}
        \STATE $\beta_k = \frac{\floor{k/\kappa}}{\floor{k/\kappa}+1}$ 
        \STATE On a random batch of transitions $B_{q,k} \subset \mathcal{B}$, update $\theta$ with one step of SGD of $\mathcal{L}_q$, see~\eqref{eq:qtarget} \label{line:lq}
        \STATE On a random batch of transitions $B_{h,k} \subset \mathcal{B}$,
            update $\phi$ with one step of SGD of  $\mathcal{L}_h$, see~\eqref{eq:htarget} \label{line:h}
    \ENDIF
    \IF{$k \mod C == 0$}
        \STATE $\old\theta \leftarrow \theta$, $\old\phi \leftarrow \phi$
    \ENDIF
\ENDFOR
\RETURN $\gr(H_\phi)$
\end{algorithmic}

\end{algorithm}

\section{Related work and discussion}
\label{sec:related}
The closest approaches to MoVI are Speedy Q-Learning (SQL) \citep{ghavamzadeh2011speedy} and Dynamic Policy Programming (DPP) \citep{azar2012dynamic} (generalized by~\citet{kozuno2019theoretical} as Conservative VI, with similar guarantees). Both approaches are extensions of AVI that also benefit from a similar compensation of errors along iterations. As fat as we know, they are the sole algorithms with this kind of guarantee. We first discuss extensively the links to SQL and DPP, before mentioning other (less) related works.

\paragraph{Algorithmic comparison.}
First, let us consider DPP, in the DPP-RL version\footnote{DPP considers general softmax policies, of which greedy policies are a special case, that correspond to DPP-RL.} \citep[Algorithm 2]{azar2012dynamic}. Define the scalar product on $\actions$ for all policy $\pi$ and $q$-value $q$ as $\langle \pi, q \rangle (s) = \sum_{a\in\actions} \pi(a|s)q(s,a)$. DPP estimates a quantity $\psi_k\in\mathbb{R}^{\states\times\actions}$, as
\begin{equation}
\label{eq:psi-dpp}
  \psi_k = \psi_{k-1} + T_{\pi_k} \psi_{k-1} - \langle \pi_k, \psi_{k-1} \rangle + \epsilon_k,
\end{equation}
with $\pi_k \in \gr(\psi_k)$. Without error, $\psi_k(s,a)$ converges to $q_*(s,a)$ when $a$ is the optimal action in state $s$, and to $-\infty$ otherwise. This makes difficult an extension of DPP to a function approximation setting (unbounded function).

Secondly, SQL updates a $q$-value $q_k$ as %
\begin{multline}
    q_k = q_{k-1} + \frac{1}{k}(T_*q_{k-2} - q_{k-1}) \\
        + \frac{k-1}{k}(T_*q_{k-1} - T_*q_{k-2}).
        \label{eq:sql-q}
\end{multline}
We then re-write SQL as an update on similar quantities as DPP. Let us define $\psi_k = k q_k$, and consider the policy $\pi_k = \gr(q_k) = \gr(\psi_k)$. SQL is then equivalent to
\begin{equation}
\label{eq:psi-sql}
    \psi_k = \psi_{k-1} + T_{\pi_k} \psi_{k-1} - \gamma P_{\pi_{k-1}} \psi_{k-2} + \epsilon_k.
\end{equation}

Finally, we also position MoVI in this setting. Here, we define $\psi_k$ as $\psi_k = (k +1)h_k = \sum_{i=0}^{k}q_j$. We consider the sequence of policies $\pi_k = \gr(h_k) = \gr(\psi_k)$. With some work (detailed in Appendix) we can rewrite Equation~\eqref{eq:movi} as an update on $\psi_k$ as
\begin{equation}
\label{eq:psi-movi}
    \psi_k = \psi_{k-1} + T_{\pi_k} \psi_{k-1} - \gamma P_{\pi_k} \psi_{k-2} + \epsilon_k.
\end{equation}
Comparing MoVI, SQL and DPP through the prism of Eqs.~\eqref{eq:psi-movi}, \eqref{eq:psi-dpp} and~\eqref{eq:psi-sql}, we observe that these three schemes are similar. They all share the first part of their update in common, and differ only in the subtraction term -- that allows for error compensation. This term is $\gamma P_{\pi_k}\psi_{k-2}$ in MoVI, which is replaced by a $\gamma P_{\pi_{k-1}}\psi_{k-2} = T_* \psi_{k-2}$ in SQL, and by $\langle \pi_k, \psi_{k-1} \rangle$ in DPP. This writing eases comparison, but we highlight that it is not how algorithms are defined initially, and implemented, except for DPP (SQL and MoVI do not require estimating an unbounded function).

\paragraph{Performance bounds.}
We now compare performance bounds of various algorithm. SQL and DPP both propagates averaged errors instead of errors, as they both satisfy\footnote{The bounds in the original papers differ slightly by their multiplicative constants, the one provided here is true for both.}
\begin{equation}
    \Vert q_* - q_{\pi_{k}} \Vert_{\infty} \leq \frac{2\gamma}{k(1-\gamma)} \Bigg( \sum_{j=1}^{k} \gamma^{k-j} \Vert {E}_{j}  \Vert_{\infty}  + \frac{8\gamma q_{\max}}{1 - \gamma} \Bigg),
\end{equation}

This is to be compared to the bound for MoVI given in Eq.~\eqref{eq:supremum}. MoVI, DPP and SQl enjoys similar bounds, the main difference being in the nature of the error terms. Both SQL and DPP depend on a term of the from $\sum_{j=0}^{k}\gamma^{k-j}\norm{E_{k}}_\infty$, that is a discounted sum of the norm of averaged errors. On the other hand, in MoVI, we have the dependency in $\sum_{j=0}^{k}\gamma^{k-j}\norm{\p{E}_{k,j}}_\infty$, so not averaged errors, but averaged weighted errors. In the generative model setting, the bound is less favourable for MoVI. Indeed, in this case, the errors are zero-mean, so the dependence on their average in DPP and SQL is a strong advantage. However, we empirically show that MoVI behaves similarly to SQL and DPP in this case (see Sec.~\ref{sec:garnets}). 
In a more general case ($\epsilon_k$ corresponding to a regression error), none of the bounds can be easily instantiated, because the quantity we can hope to control is $\|\epsilon_k\|_2,\mu$, not $\|E_k\|_{2,\mu}$. This, we will check the algorithms' behaviors empirically.

From a practical point of view, neither SQL or DPP have been originally implemented in RL on large scale problems. A deep version of a variation of DPP\footnote{Specifically, it is the update described by~\citet[Eq.~(24)]{azar2012dynamic}, that also lead to an asymptotically unbounded function, and thus to numerical instability.} have been proposed by \citet{tsurumine2017deep}, but it is only applied on a small number of samples. The principal issue of a practical DPP is that it has to estimate $\psi_k$, a quantity that is asymptotically unbounded. It could then be applied on short training environments, when this value is updated a relatively small number of times, and stays numerically stable. However, on environments like the ALE, where one needs to compute millions of environments steps, DPP is likely to diverge, and fail due to numerical issues. In Section~\ref{sec:atari}, we provide a experiment in a larger setting. We extened MoVI to deep learning, and, for the sake of somparison, we propose deep versions of SQL and DPP. These two last algorithms are variations of DQN that make use of updates in Equations~\eqref{eq:sql-q} and~\eqref{eq:psi-dpp} to define DQN-like regression targets. We could not obtain satisfying results with both of these implementations. Experimental results and details are given in Section~\eqref{sec:experiments} and in the Appendix.

\paragraph{Other related methods.}
MoVI shares also algorithmic similarities with other algorithms, Softened LSPI \citep{perolat2016softened} and Politex \citep{lazic2019politex}. \citet{perolat2016softened} consider the zero-sum games setting, and propose a Policy Iteration (PI)-based algorithm. It relates to MoVI in the sense that it averages the $q$-values of consecutive policies. Politex is also a PI-scheme, where the policy is a softmax of the sum of all $q$-values. These two algorithms share the idea of averaging the $q$-values, but are derived from different principles. \citet{perolat2016softened} build their algorithm as a quasi-Newton method on the Bellman residual and rely heavily on linear parameterization, while Politex build upon prediction with expert advice, and deals with the average reward criterion, instead of the discounted one. Moreover, none of these two approaches offer the kind of guarantee about the propagation of averaged errors that DPP, SQL or MoVI have.

\section{Experiments}
\label{sec:experiments}
In this Section, we present experimental results from MoVI and Momentum-DQN. 
First, we consider small random MDPs (Garnets), to check empirically Asm.~\ref{co:filtration} and to compare to DDP and SQL on a tabular setting, with access to a generative model. Then, we experiment Momentum-DQN on a subset of Atari games, and compare to DQN (a natural baseline) as well as deep versions of DPP and SQL. Further experimental details are provided in the appendix.

\subsection{Garnets}
\label{sec:garnets}
A Garnet~\citep{archibald1995generation,bhatnagar2009natural} is an abstract MDP. It is built from three parameters ($N_S$, $N_A$, $N_B$). $N_S$ and $N_A$ are respectively the number of states and actions. The parameter $N_B$ is the branching factor, the maximum number of states accessible from any other state. The transition probabilities $P(\p s | s, a)$ are then computed as follows. For each state-action couple $(s,a)$, $N_B$ states ($s_1, \hdots s_{N_B}$) are drawn uniformly without replacement. Then, $N_B - 1$ number are drawn uniformly in $(0,1)$ and sorted as $(p_0=0, p_1, \hdots p_{N_B-1}, p_{N_B}=1)$. The transition probabilities are assigned as $P(s_k|s,a)=p_{k} - p_{k-1}$ for each $1\leq k \leq N_B$. The reward function is drawn uniformly in $(-1,1)^{N_S}$.

\begin{figure}
    \centering
    \includegraphics[width=\linewidth]{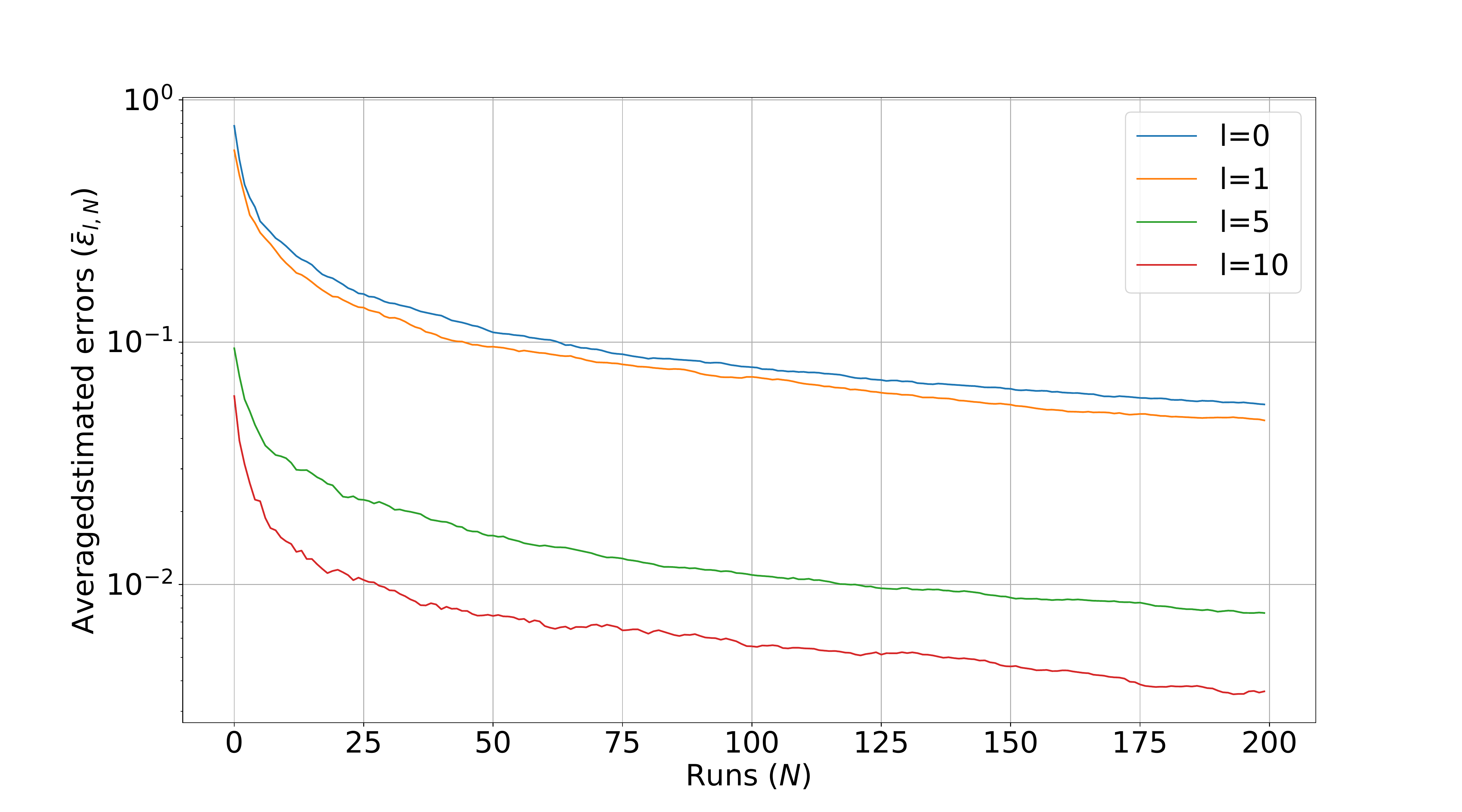}
    \caption{Evolution of the empirical weighted average error $\bar\epsilon_{l,N}$ with $N$ (log scale) for different values of $l$. We need a convergence towards $0$ for our assumption to be numerically verified, which seems to be the case. }
    \label{fig:errors}
\end{figure}

\paragraph{Assumption check.}
First, we want to check that Asm.~\ref{co:filtration} is reasonable. Given a step $j$ of the algorithm and a size $l$, we compute an empirical estimate of $\mathbb{E}[P_{j+l:j+1}\epsilon_j]$. With Garnets, we have access to the transition kernel, so we can compute the error at step $j$, $\epsilon_j(s,a) = \hat{T}_{\pi_j} q_j(s,a) - T_{\pi_j} q_j(s,a)$. Given a fixed Garnet, we first compute the value $q_j$ with MoVI. Then, on a number $N$ of runs, we re-start MoVI from the same $q_j$, re-run the algorithm for $l$ steps from there, and compute the values $P_{j+l:j+1, n} \epsilon_{j,n}(s,a)$, with $n \in [|1;N|]$. We get an estimate $\bar\epsilon_{l, N}$ of $\norm{\mathbb{E}[P_{j+l:j+1}\epsilon_j|\mathcal{F}_{j-1}]}_\infty$,
\begin{equation}
    \bar\epsilon_{l,N} = \max_{(s,a) \in \states\times\actions} \abs{\frac{1}{N} \sum_{n=1}^N P_{j+l:j+1, n} \epsilon_{j,n}(s,a)}.
\end{equation}
We want to check that $\bar\epsilon_N \rightarrow 0$ when $N \rightarrow \infty$. For several values of $l$, We compute $\bar\epsilon_{l,N}$ for $N$ between $0$ and $200$, and average these results over $100$ garnets. We report the evolution of the means of $\bar\epsilon_{l,N}$ (over Garnets) in Figure~\ref{fig:errors}. We observe that the limit of $\bar\epsilon_{l,N}$ seems to be $0$ for each $l$, which experimentally validates our assumption. With $l=0$, we get the ``natural'' norm of errors (not multiplied by any matrix). We see here that, for every tested $l>0$, the norm is lower than for $l=0$, meaning that the policies kernels do not have a negative impact on the expected value, but seem to further reduce variance.

\begin{figure}
    \centering
    \includegraphics[width=\linewidth]{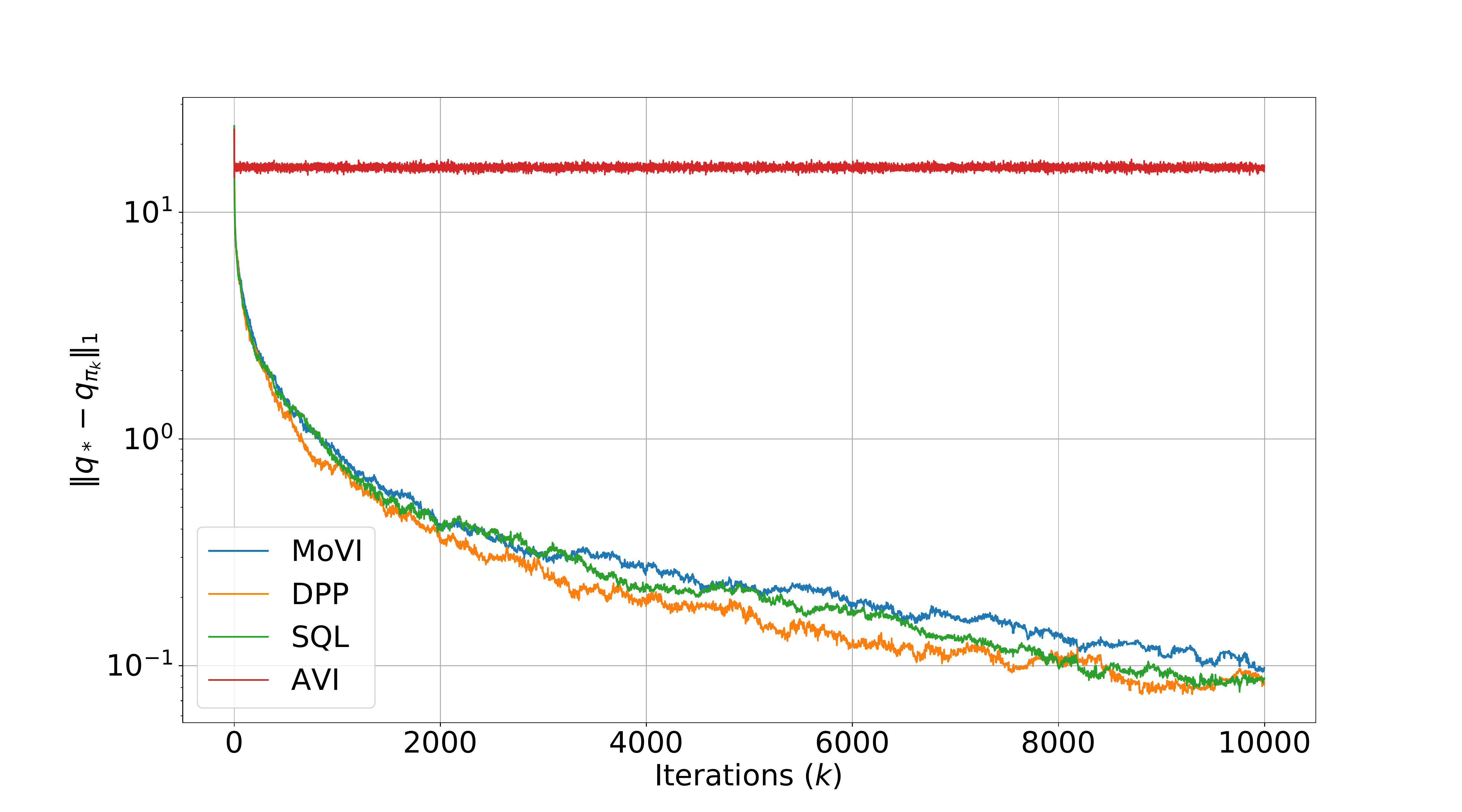}
    \caption{Error on the policy value of different ADP schemes. Each curve represents $\norm{q_* - q_{\pi_k}}_1$, where $\pi_k$ results from AVI, MoVI, SQL or DPP.}%
    \label{fig:adp}
\end{figure}

\paragraph{Algorithms comparison.}
We compare VI, MoVI, SQL and DPP on random Garnets, using the sampled version with a generative model described in Section~\ref{sec:sample}. We run each algorithm on $100$ Garnets, an we report the norm of the empirical error on  the uniform distribution $\norm{q_* - q_{\pi_k}}_1$. We can compute the exact value of $\pi_k$ with access to the model. The four algorithms are compared in Figure~\ref{fig:adp}. We observe an almost identical behaviour for MoVI, DPP, and SQL. They all converge towards $v_*$ at roughly the same speed, while AVI oscillates around a sub-optimal policy.

\subsection{Atari}
\label{sec:atari}
Atari is a standard discrete-actions environment introduced by~\cite{bellemare2013arcade} with a high dimensional state space. We use this environment to validate our Momentum-DQN architecture. Our baseline is DQN as it is implemented in the Dopamine library~\citep{castro2018dopamine}. We used the same architecture and the same hyperparameters as DQN, and notably we used sticky action with a rate of $0.25$ to introduce stochasticity as recommended by \citet{machado2018revisiting}, and our state consists in the stacking of the $4$ last frames. Every $4$ steps in the environment, we perform a gradient update on $\theta$ and $\phi$. Every C=25000 environment steps, we update the target networks. We report the average undiscounted score obtained during learning on the last $250000$ steps (named an iteration). On the figures, the thick line show this average score averaged on $5$ random seeds, while the semi-transparent parts denote the standard deviation with respect to the seeds.

\begin{figure}[tbh]
    \centering
    \includegraphics[width=\linewidth]{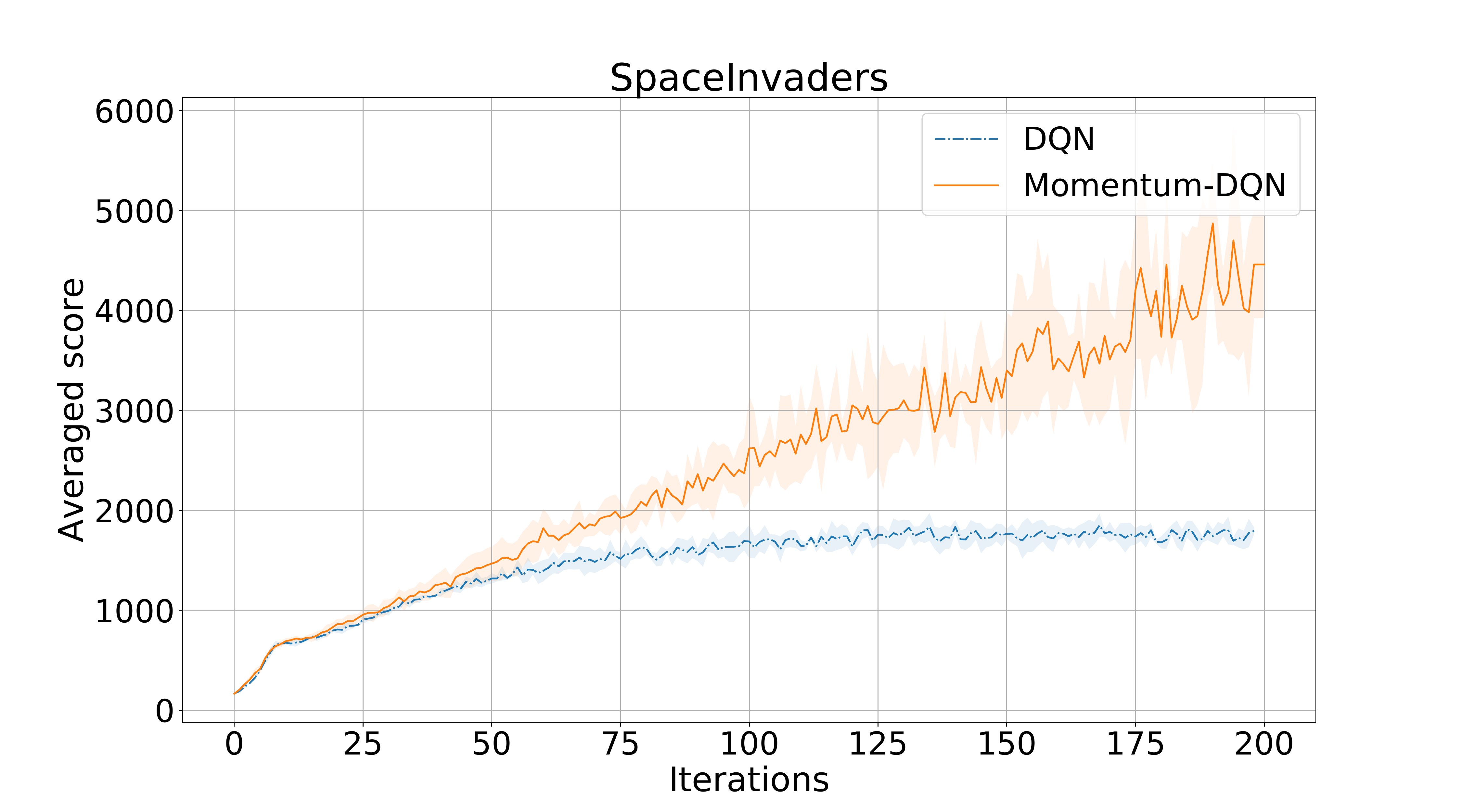}
    \caption{Scores obtained on SpaceInvaders by DQN (dashed-dotted, blue) and Momentum-DQN (orange).} 
    \label{fig:sea}
\end{figure}

\begin{figure}[tbh]
    \centering
    \includegraphics[width=\linewidth]{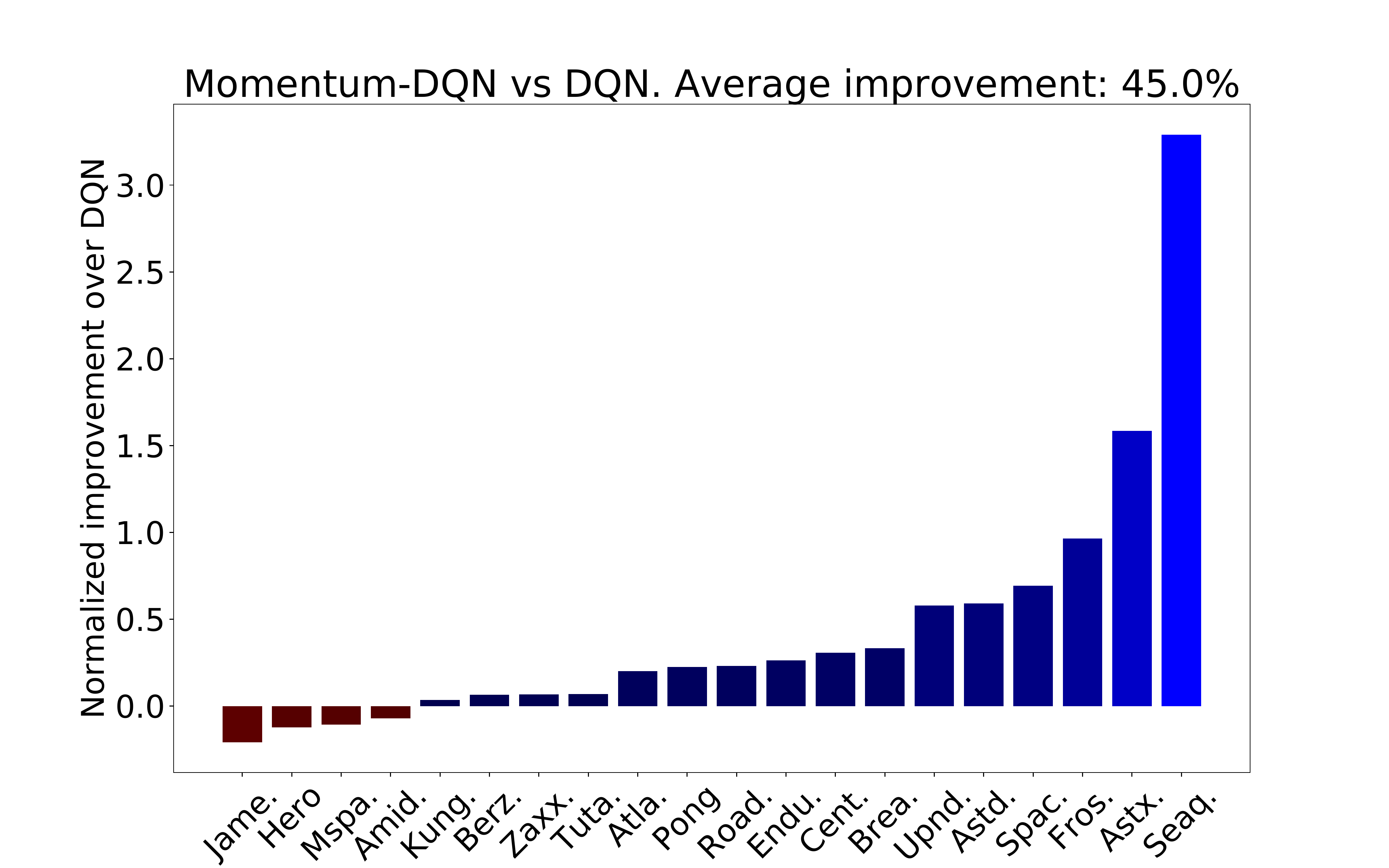}
    \caption{Normalized improvement of Momentum-DQN over DQN. We obtain an almost constant improvement on these $20$ games.} 
    \label{fig:auc}
\end{figure}

We evaluate Momentum-DQN on a subset of $20$ Atari games. This games are selected to represent the categories from~\citet[Appendix A]{ostrovski2017count}, excluding the hardest exploration ones -- we have no claim in helping DQN in this setting. Here, we used a schedule of $\beta_k$ as defined in Section ~\ref{sec:modqn}, with $\kappa=2500000$ that we tuned on a small subset of game (Asterix, Zaxxon, and Jamesbond). As an example, we give the comparison of Momentum-DQN and DQN on the game SpaceInvaders in Figure~\ref{fig:sea}. In figure~\ref{fig:auc}, we report the normalized improvement of Momentum-DQN over DQN using the Area Under the Curve (AUC) metric. These results show a clear improvement using Momentum. Momentum-DQN outperforms DQN on $16$ games out of $20$, with an average normalized improvement of $45\%$. It only under-performs DQN on three games by a low margin, while the improvement goes up to $200\%$ for the game Seaquest. In the Appendix, we report the score obtained for the $20$ games, along with experiments testing the influence of various $\beta_k$ schedules.

\begin{figure}[tbh]
    \centering
    \includegraphics[width=\linewidth]{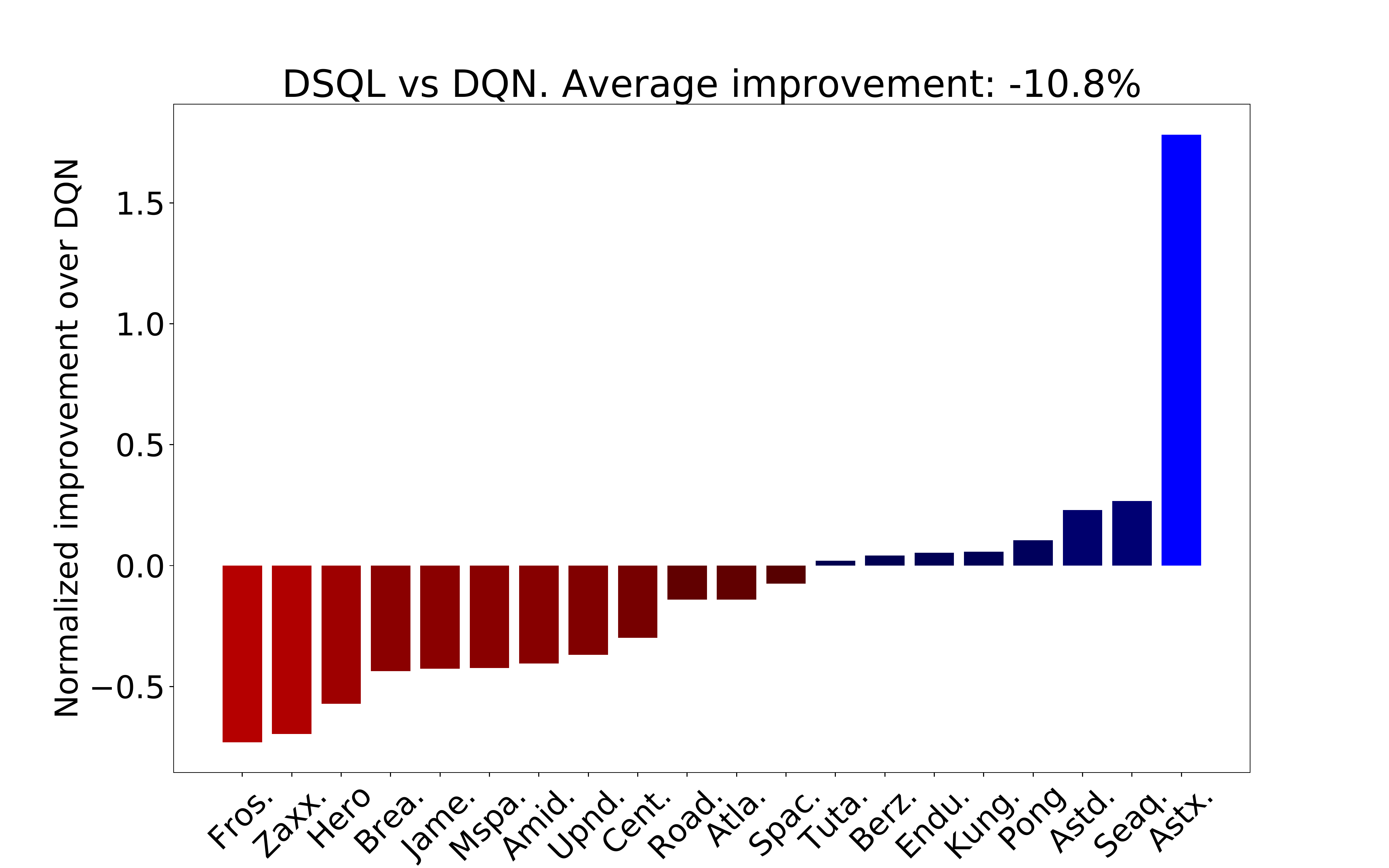}
    \caption{Normalized improvement of DSQL over DQN.} 
    \label{fig:auc-sql}
\end{figure}

\begin{figure}[tbh]
    \centering
    \includegraphics[width=\linewidth]{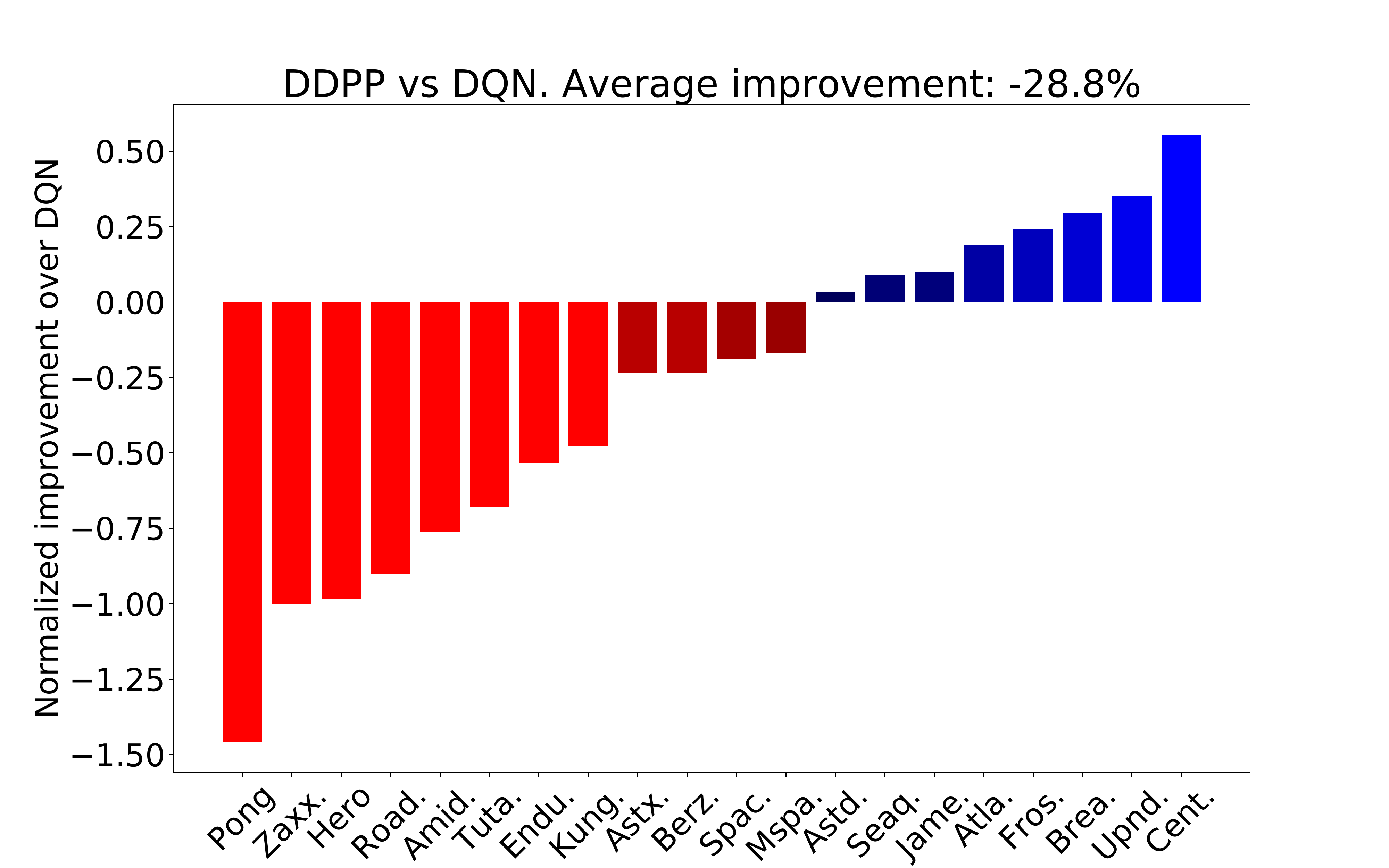}
    \caption{Normalized improvement of DDPP over DQN.} 
    \label{fig:auc-dpp}
\end{figure}

\paragraph{Deep-SQL and Deep-DPP.}
We implemented Deep versions of SQL and DPP (respectively DSQL and DDPP), that we tested on Atari, also based on the architecture and hyperparameters of Dopamine's DQN. For both algorithms, we derive an update rule based on the ADP scheme, using the same parametrization as DQN (we report specific equations in Appendix). We were however not able to obtain satisfying -- i.e. competitive with DQN -- scores with these algorithms. We report the experimental results of DDPP and DSQL versus DQN in Figures~\ref{fig:auc-sql} and~\ref{fig:auc-dpp}. We used the same parameters as for Momentum-DQN, in particular the same $\beta_k$ schedule for DSQL. On these two graphs, we see that both DSQL and DDPP underperform DQN on most of the games. 

For DDPP, the reason is quite simple, as the $Q$-network has to estimate a value that diverges to $-\infty$, causing heavy numerical issues, and the algorithms fails on most of the games after a few iterations. It is less clear why DSQL underperforms DQN. Our hypothesis is that Momentum-DQN enjoys a separate network that approximate the average of the $q$-values, while DSQL needs to compute its update from consecutive target. However, when using deep networks and stochastic approximation, the consecutive target networks cannot securely be associated to consecutive $q$-values computed in ADP, making the update in DSQL less reliable.  

\section{Conclusion}
We introduced a new ADP scheme, MoVI, inspired by Momentum in gradient ascent. To adapt Momentum to RL, we made an analogy between the $q$-values in DP schemes and the gradient in gradient ascent methods, interpreting Momentum in RL as an averaging of consecutive $q$-function. We provided an anlysis of MoVI, showing that the Momentum brings compensation of errors to AVI. We also derived a partial analysis of the sample complexity when instantiated in the tabular case. 
These results are similar to what are to our knowledge the closest algorithms to MoVI, SQL and DPP. Our bound involves a more complicated averaging of errors, extensively discussed. Yet, we have shown that all algorithmic schemes behave similarly in toy problems.
We advocated that MoVI is better suited for deep learning extensions and proposed Momentum-DQN, as well as natural deep extensions of DPP and SQL. With experiments on a representative subset of Atari games, we have shown that, contrary to DDPP and DSQL, momentum-DQN brings a clear improvement over DQN.
Note that in principle, Momentum could be applied to any RL algorithm that estimates a value: a value-based algorithm like C51~\citep{bellemare2017distributional}, or an actor-critic (for example, SAC~\citep{haarnoja2018soft}). It could also be extended straightforwardly to continuous action settings, replacing the critic by the average of successive critics. We plan to extend the idea of Momentum to other RL algorithms in future works.

\clearpage

\bibliographystyle{plainnat}
\bibliography{biblio}

\iftrue

\clearpage
\onecolumn
\appendix

\section*{Content}
This Appendix provides the proofs of all results stated in the paper, along with additional experiments and experimental details. 

In Appendix~\ref{app:proof-thm}, we give the proof of Theorem~\ref{th:bound}, the proof of corollary~\ref{th:bound-1mu} is given in Appendix~\ref{app:proof-cor}, and the proof of proposition~\ref{th:sc} is given in Appendix~\ref{app:proof-prop}. In Appendix~\ref{app:comp}, we give details on how we could express MoVI and SQL in their formulations of Equations~\eqref{eq:psi-movi} and~\eqref{eq:psi-sql}.

Then, we complete the experiments presented in the paper. We give additional details and complete results on Garnets in Appendix~\ref{app:garnets}. Appendix~\ref{app:atari} completes the experiments on Atari. The detailed update equations of DSQL and DDPP are given in Appendix~\ref{app:atari-dsql}. In Appendix~\ref{app:atari-exp}, we provide details about how we conducted experiments, in Appendix~\ref{app:atari-mixt} we discuss the influence of the mixture rate, and finally we present full results (on the $20$ games) in Appendix~\ref{app:atari-joint} and~\ref{app:atari-full}.

\section{Proof of Theorem~\ref{th:bound} \label{app:proof-thm}}
In this Section, we give the proof of Theorem~\ref{th:bound}. Let us recall the definition of MoVI in this case,
\begin{equation}
\label{eq:movi-app}
    \begin{cases}
    \pi_{k} = \gr(h_{k-1}) \\
    q_{k} = T_{\pi_{k}} q_{k-1} + \epsilon_{k}  \\
    h_{k} = \frac{k}{k+1} h_{k-1} + \frac{1}{k+1} q_{k} .
    \end{cases}
\end{equation}

We want to prove the following component-wise inequality (Theorem~\ref{th:bound})
\begin{equation}
    q_* - q_{\pi_{k+1}}  \leq \frac{1}{k+1} \Bigg[ (I - \gamma P_{\pi_*})^{-1} (E_{k+1} + q_{k+1} - q_0)
                         - (I - \gamma P_{\pi_{k+1}})^{-1} \bigg(\sum_{j=0}^{k-1} \gamma^j\p{E}_{k,j}
                            +\sum_{j=0}^k \gamma^j P_{j:1}(T_{\pi_1} q_0 - q_0)\bigg) \Bigg].
\end{equation}

First, we prove a useful lemma, that essentially tells that controling a residual is enough.
\begin{lemma}
\label{lemm:qpi}
For any $\pi$ and $q$, we have
\begin{equation}
    q_\pi - q = (I-\gamma P_\pi)^{-1}(T_\pi q - q).
\end{equation}
\end{lemma}
\begin{proof}
\begin{align}
    q_\pi - q &= T_\pi q_\pi - T_\pi q + T_\pi q - q
    \\
    &= \gamma P_\pi(q_\pi - q) + T_\pi q - q
    \\
    \Leftrightarrow
    q_\pi - q &= (I-\gamma P_\pi)^{-1}(T_\pi q - q).
\end{align}
\end{proof}

Now, let us get to the central proof. We use the following decomposition
\begin{equation}
    q_* - q_{\pi_{k+1}} = q_* - h_k + h_k - q_{\pi_{k+1}}.
\end{equation}
Applying Lemma~\ref{lemm:qpi} to $q_* - h_k$ and $q_{\pi_{k+1}} - h_k$, we have that
\begin{equation}
    q_* - q_{\pi_{k+1}} = (I-\gamma P_{\pi_*})^{-1} (T_{\pi_*} h_k - h_k) -  (I-\gamma P_{\pi_{k+1}})^{-1}(T_{\pi_{k+1}} h_k - h_k).
\end{equation}
Using the fact that $T_{\pi_*} h_k \leq T_{\pi_{k+1}} h_k$ (as $\pi_{k+1}  = \gr(h_k)$), we have
\begin{equation}
\label{eq:decomp-bound}
    q_* - q_{\pi_{k+1}} \leq (I-\gamma P_{\pi_*})^{-1} (T_{\pi_k+1} h_k - h_k) -  (I-\gamma P_{\pi_{k+1}})^{-1}(T_{\pi_{k+1}} h_k - h_k).
\end{equation}
From here, we then only need to upper bound and lower bound the residual $T_{\pi_k+1} h_k - h_k$.

\subsection{Upper bound of the residual}

We have, using the update definition, and the fact that $\pi_k \in \gr(h_{k-1})$
\begin{align}
    T_{\pi_{k+1}} h_k &= \frac{k}{k+1} T_{\pi_{k+1}} h_{k-1} + \frac{1}{k+1} T_{\pi_{k+1}} q_k
    \\
    &= \frac{k}{k+1} T_{\pi_{k}} h_{k-1} + \frac{1}{k+1}(q_{k+1} - \epsilon_{k+1})\\
    \Rightarrow (k+1) T_{\pi_{k+1}} h_k &\leq k T_{\pi_{k}} h_{k-1} + q_{k+1} -\epsilon_{k+1}.
\end{align}
By direct induction, we obtain
\begin{align}
    (k+1) T_{\pi_{k+1}} h_k &\leq k T_{\pi_{k}} h_{k-1} + q_{k+1} -\epsilon_{k+1}
    \\
    &\leq \sum_{j=1}^{k+1} q_j - \sum_{j=1}^{k+1} \epsilon_j. \label{eq:ub-0}
\end{align}
By definition of $h_k$, $h_{k} = \frac{k}{k+1} h_{k-1} + \frac{1}{k+1} q_{k} = \frac{1}{k+1}\sum_{j=0}^k q_k$, thus
\begin{equation}
\label{eq:cumul_err}
    \sum_{j=1}^{k+1} q_j = \sum_{j=0}^{k} q_j  + q_{k+1} - q_0 = (k+1) h_k + q_{k+1} - q_0.
\end{equation}
We define the negative cumulative error $E_{k+1} = -\sum_{j=1}^{k+1}\epsilon_j$. Using this definition and Equation~\eqref{eq:cumul_err} in Equation~\eqref{eq:ub-0}, we have
\begin{align}
\label{eq:upper-bound}
    T_{\pi_{k+1}} h_k - h_k &\leq \frac{1}{k+1}(q_{k+1} - q_0 + E_{k+1}),
\end{align}
an upper bound on the residual.

\subsection{Lower bound of the residual}
\label{subsec:notnice}

Using the definition of $h_k$ and $\pi_{k+1}\in\gr(h_k)$, and using an induction argument, we have
\begin{align}
    (k+1) T_{\pi_{k+1}} h_k &\geq (k+1) T_{\pi_k} h_k 
    \\
    &= k T_{\pi_k} h_{k-1} + T_{\pi_k} q_k
    \\
    &\geq \sum_{j=1}^k T_{\pi_j} q_j + T_{\pi_1} q_0. \label{eq:lb-0}
\end{align}
Using this, we can then lower bound the residual by a sum of others residuals:
\begin{align}
    (k+1) (T_{\pi_{k+1}} h_k - h_k) &\geq \sum_{j=1}^k T_{\pi_j} q_j + T_{\pi_1} q_0 - \sum_{j=0}^k q_j
    \\
    &= \sum_{j=1}^k (T_{\pi_j} q_j - q_j) +  T_{\pi_1} q_0  - q_0.
\end{align}

Let work on one of these residuals. We have
\begin{align}
    T_{\pi_j} q_j - q_j &= T_{\pi_j}(T_{\pi_j}q_{j-1} + \epsilon_j) - (T_{\pi_j}q_{j-1} + \epsilon_j)
    \\
    &= T_{\pi_j}^2 q_{j-1} - T_{\pi_j} q_{j-1} - (I-\gamma P_{\pi_j})\epsilon_j
    \\
    &= \gamma P_{\pi_j} (T_{\pi_j} q_{j-1} - q_{j-1}) - (I-\gamma P_{\pi_j})\epsilon_j
\end{align}
On the other side, using the fact that by definition $q_k = (k+1) h_k - k h_{k-1}$, we have that
\begin{align}
    T_{\pi_{j+1}} q_{j} - T_{\pi_{j}} q_{j} &= (j+1) T_{\pi_{j+1}} h_j - j T_{\pi_{j+1}} h_{j-1} - (j+1) T_{\pi_{j}} h_j + j T_{\pi_j} h_{j-1}
    \\
    &= (j+1)(T_{\pi_{j+1}} h_j - T_{\pi_{j}} h_j) + j(T_{\pi_j} h_{j-1} - T_{\pi_{j+1}} h_{j-1})
    \\
    &\geq 0.
\end{align}

Therefore, we can conclude that
\begin{align}
    T_{\pi_j} q_j - q_j &= \gamma P_{\pi_j} (T_{\pi_j} q_{j-1} - q_{j-1}) - (I-\gamma P_{\pi_j})\epsilon_j
    \\
    &\geq \gamma P_{\pi_j} (T_{\pi_{j-1}} q_{j-1} - q_{j-1}) - (I-\gamma P_{\pi_j})\epsilon_j
\end{align}

Write $P_{j:i} = P_{\pi_j} P_{\pi_{j-1}} \dots P_{\pi_i}$ if $1\leq i \leq j$, $P_{j:i} = I$ otherwise. We have by induction
\begin{equation}
    T_{\pi_j} q_j - q_j \geq - \sum_{i=1}^{j} \gamma^{j-i} P_{j:i+1}(I-\gamma P_{\pi_i}) \epsilon_i + \gamma^{j+1} P_{j:1}(T_{\pi_1} q_0 - q_0).
\end{equation}

Plugging this in the inequality from Equation~\eqref{eq:lb-0}, we get
\begin{align}
    (k+1)(T_{\pi_{k+1}} h_k - h_k) &\geq  \sum_{j=1}^k (T_{\pi_j} q_j-q_j) + T_{\pi_1}q_0 - q_0 \\
    &\geq \sum_{j=1}^k \left( -\sum_{i=1}^{j} \gamma^{j-i} P_{j:i+1} (I - \gamma P_{\pi_i})\epsilon_i + \gamma^{j+1} P_{j:1} (T_{\pi_1}q_0 - q_0) \right) + T_{\pi_1}q_0 - q_0 \\
    &\geq -\sum_{j=0}^{k-1} \gamma^j \sum_{i=1}^{k-j} P_{i+j:i+1} (I - \gamma P_{\pi_i})\epsilon_i + \sum_{j=0}^k \gamma^{j} P_{j:1} (T_{\pi_1}q_0 - q_0).
\end{align}

We define the weighted error $\p{E}_{k,j} = -\sum_{i = 1}^{k-j} P_{i+j:i+1}(I - \gamma P_{{\pi}_i}) \epsilon_i$. With this definition, we have that
\begin{equation}
    \label{eq:lower-bound}
    T_{\pi_{k+1}} h_k - h_k \geq \frac{1}{k+1} \left(
     \sum_{j=0}^k \gamma^{j} P_{j:1} (T_{\pi_1}q_0 - q_0) + \sum_{j=0}^{k-1} \gamma^j  \p{E}_{k,j}
     \right),
\end{equation}
a lower bound on the residual.

Using the lower bound from Equation~\eqref{eq:lower-bound} and the upper bound from Equation~\eqref{eq:upper-bound} into the decomposition of Equation~\eqref{eq:decomp-bound} proves Theorem~\ref{th:bound}. \begin{flushright}$\square$\end{flushright}

\section{Proof of Corollary~\ref{th:bound-1mu} \label{app:proof-cor}}
We use the previous result to prove corollary~\ref{th:bound-1mu}, the error propagation in $\mu$-weighted $\ell_1$-norm. Let $\mu$ be the distribution of interest (where we want to control the error), and $\nu$ the sampling distribution (from where we have access to transitions). We have, directly from Theorem~\ref{th:bound} (using the fact that $\mu (I-\gamma P_\pi)^{-1} = (1-\gamma)^{-1} d_{\pi,\mu}$, and noticing that $q_* - q_{\pi_k} \geq 0$),
\begin{multline}
\label{bound-1mu-0}
    \norm{q_* - q_{\pi_{k+1}}}_{1,\mu}  \leq \underbrace{\frac{1}{(k+1)(1-\gamma)} \left(d_{\pi_*,\mu} \abs{E_{k+1}} + d_{\pi_{k+1},\mu} \sum_{j=0}^{k-1}\gamma^j\abs{\p{E}_{k,j}}  \right)}_{A_k} \\
    + \underbrace{\frac{\norm{q_{k+1}}_\infty + \norm{q_0}_\infty}{(k+1)(1 - \gamma)}  + \frac{1}{(k+1)(1-\gamma)} \sum_{j=0}^{k}\gamma^j\norm{T_{\pi_1}q_0 - q_0}_\infty}_{B_k}.
\end{multline}
Let us work on the term $A_k$. We have, using the fact that $d_{\pi,\mu} \leq \|\frac{d_{\pi,\mu}}{\nu}\|_\infty \nu$, that
\begin{align}
    A_k &= \frac{1}{(k+1)(1-\gamma)} \left(d_{\pi_*,\mu} \abs{E_{k+1}} + d_{\pi_{k+1},\mu} \sum_{j=0}^{k-1}\gamma^j\abs{\p{E}_{k,j}} \right)\\
        &\leq \frac{1}{(k+1)(1-\gamma)} \left(\norm{\frac{d_{\pi_*, \mu}}{\nu}}_\infty \nu\abs{E_{k+1}} + \norm{\frac{d_{\pi_{k+1},\mu}}{\nu}}_\infty \nu\sum_{j=0}^{k-1}\gamma^j\abs{\p{E}_{k,j}} \right). \label{eq:bound-1mu-ak-0}
\end{align}
We now introduce the following concentrability coefficient
\begin{equation}
 C = \max_{\pi} \norm{\frac{d_{\pi, \mu}}{\nu}}_\infty,
\end{equation}
and we have directly from Equation~\eqref{eq:bound-1mu-ak-0} that
\begin{equation}
    A_k \leq \frac{C}{(k+1)(1-\gamma)} \left(\norm{E_{k+1}}_{1,\nu} + \sum_{j=0}^{k-1}\gamma^j\norm{\p{E}_{k,j}}_{1,\nu} \right). \label{bound-ak}
\end{equation}

Now, we upper bound the term $B_k$. Assume that we initialize MoVI with $h_0= q_0 = 0$. We have
\begin{align}
    B_k &= \frac{\norm{q_{k+1}}_\infty}{(k+1)(1 - \gamma)}  + \frac{1}{(k+1)(1-\gamma)} \sum_{j=0}^{k}\gamma^j\underbrace{\norm{T_{\pi_1}0}_\infty}_{\leq r_{\max}} \\
        &\leq \frac{q_{\max}}{(k+1)(1 - \gamma)} + \frac{(1- \gamma^{k+1})r_{\max}}{(k+1)(1-\gamma)^2} \\
        &\leq \frac{2q_{\max}}{(k+1)(1-\gamma)}, \label{bound-bk}
\end{align}
where we used $q_{\max} = r_{\max}/(1 - \gamma)$.

Using Equations~\eqref{bound-ak} and~\eqref{bound-bk} in Equation~\eqref{bound-1mu-0} proves Corollary~\ref{th:bound-1mu}. \begin{flushright}$\square$\end{flushright}

\section{Proof of Proposition~\ref{th:sc} \label{app:proof-prop}}
In this section, we prove our result in sample complexity. Essentially, we use the same method as~\citet{ghavamzadeh2011speedy}. Note that we need to have Asm.~\ref{co:filtration} for the proof method to work on our case.

Let us recall the result in supremum norm
\begin{equation}
     \Vert q_* - q_{\pi_{k+1}} \Vert_{\infty} \leq \frac{1}{(k+1)(1-\gamma)} \Bigg( \Vert E_{k+1} \Vert_{\infty} +
											  \sum_{j=0}^{k-1} \gamma^j \Vert \p{E}_{k,j} \Vert_{\infty} + 2q_{\max} \Bigg).
\end{equation}
We want to prove that, with probability at least $1-\delta$,
\begin{equation}
    \Vert q_* - q_{\pi_{k+1}} \Vert_\infty \leq \frac{2r_{\max}}{(1-\gamma)^2} \Bigg[\frac{1}{k+1} + \frac{3}{(1-\gamma)}
                                                                                    \sqrt{ \frac{2 \ln{(\frac{4|\states||\actions|}
                                                                                                             {\delta})}{}
                                                                                                }{k+1}}
                                                                  \Bigg]. \quad \text{(Proposition~\ref{th:sc})}
\end{equation}
To prove the Proposition~\ref{th:sc} on the sample complexity, we will apply the same proof technique as the one used by~\citet{ghavamzadeh2011speedy} to prove their Theorem~1.

We first notice that
\begin{equation}
\label{eq:ineq-max}
    \sum_{j=0}^{k-1} \gamma^j \Vert \p{E}_{k,j} \Vert_{\infty} \leq \frac{\max_{j\leq k-1} \norm{\p{E}_{k,j}}_\infty}{1-\gamma},
\end{equation}
so we just need to bound the terms $\norm{E_{k+1}}_\infty$ and $\max_{j\leq k-1} \norm{\p{E}_{k,j}}_{\infty}$. Precisely, we need to prove the following bound with probablility at least $1-\delta$
\begin{equation}
\label{eq:max-bound}
    \norm{E_{k+1}}_\infty + \frac{1}{1 - \gamma}\max_{0\leq j\leq k-1} \norm{\p{E}_{k,j}}_{\infty} \leq \frac{6r_{\max}}{(1- \gamma)^2} \sqrt{ 2 (k+1) \ln{\frac{4|\states||\actions|}{\delta}}}.
\end{equation}

Recall the definitions of $E_{k+1} = -\sum_{j=1}^{k+1}\epsilon_j$ and $\p{E}_{k,j} = -\sum_{i = 1}^{k-j} P_{i+j:i+1}(I - \gamma P_{{\pi}_i}) \epsilon_i$. We bound in norm the individual terms in the sums with
\begin{equation}
    \norm{\epsilon_j}_\infty \leq \norm{\hat{T}_{\pi_j}q_{j-1} - T_{\pi_j}q_j}_\infty \leq 2 q_{\max}, 
\end{equation}
and
\begin{equation}
    \norm{P_{i+j:i+1}(I - \gamma P_{{\pi}_i}) \epsilon_i}_\infty \leq \norm{I - \gamma P_{\pi_i}}_\infty 2q_{\max} \leq 4q_{\max}.
\end{equation}

Using a Maximal Azuma-Hoeffding inequality in the same manner as \citet{ghavamzadeh2011speedy} on $E_{k+1}$, we have
\begin{equation}
    \mathbb{P} \left( \norm{E_{k+1}}_{\infty} \leq q_{\max} \sqrt{8 (k+1) \ln{\frac{4|\states||\actions|}{\delta}}}\right) \geq 1- \frac{\delta}{2}.
\end{equation}

Assuming that Assumption~\ref{co:filtration} holds, we can use a similar argument on $\p{E}_{k,j}$. This give the following bound
\begin{equation}
    \mathbb{P} \left( \max_{0\leq j\leq k-1} \norm{\p{E}_{k,j}}_{\infty} \leq q_{\max} \sqrt{32 (k+1) \ln{\frac{4|\states||\actions|}{\delta}}}\right) \geq 1- \frac{\delta}{2}.
\end{equation}

Combining both results, we obtain Equation~\eqref{eq:max-bound}, and so prove the result in Proposition~\ref{th:sc}. \begin{flushright}$\qed$\end{flushright}

\section{Additional proofs on algorithmic comparison \label{app:comp}}
Here we detail how we obtain the formulations of Section~\ref{sec:related}, specifically Equations~\eqref{eq:psi-movi} and~\eqref{eq:psi-sql}.

\paragraph{MoVI.}
First, we prove a recursion on $(k +1)h_k$.
\begin{lemma} 
\label{lemm:rec}
Recursion on $kh_{k-1}$.
For each $k \geq 0$,
\begin{equation}
    (k+1) h_k = k h_{k-1} +  T_{\pi_k} [k h_{k-1}] - \gamma P_{\pi_k} [(k-1) h_{k-2}] + \epsilon_k.
\end{equation}
\end{lemma}
\begin{proof}
\begin{align}
    (k+1) h_k &= k h_{k-1} + q_k
    \\
    &= k h_{k-1} + T_{\pi_k} q_{k-1} + \epsilon_k
    \\
    &= k h_{k-1} + T_{\pi_k} \left(k h_{k-1} - (k-1) h_{k-2}\right) + \epsilon_k
    \\
    &= k h_{k-1} + (k - (k-1))r + k\gamma P_{\pi_k} h_{k-1} - (k-1)\gamma P_{\pi_k} h_{k-2} + \epsilon_k
    \\
    &= k h_{k-1} +  T_{\pi_k} [k h_{k-1}] - \gamma P_{\pi_k} [(k-1) h_{k-2}] + \epsilon_k.
\end{align}
\end{proof}

Let us write $\psi_k = (k+1)h_k = \sum_{j=0}^k q_j$. Note that $\gr(h_k) = \gr(\psi_k)$. Using this in Lemma~\ref{lemm:rec},we can write the MoVI update as
\begin{equation}
    \psi_k = \psi_{k-1} + T_{\pi_k} \psi_{k-1} - \gamma P_{\pi_k} \psi_{k-2} + \epsilon_k,
\end{equation}
which proves Equation~\eqref{eq:psi-movi}.

\paragraph{SQL.}
The SQL update is
\begin{equation}
    q_k = q_{k-1} + \frac{1}{k+1}(T_*q_{k-2} - q_{k-1}) + \frac{k}{k+1}(T_*q_{k-1} - T_*q_{k-2}).
\end{equation}
We have
\begin{align}
    (k+1) q_k &= (k+1) q_{k-1} + T_*q_{k-2} - q_{k-1} + k T_* q_{k-1} - k T_* q_{k-2} \\
              &= k q_{k-1} + kT_*q_{k-1} - (k-1) T_*q_{k-2} \label{eq:sql-update-2}.
\end{align}
Let us define  $\psi_k = (k+1)h_k$ in this case. Using that $T_* \psi_k - T_* \psi_{k-1} = (k+1)T_*q_k - kT_*q_{k-1}$ and writing $\pi_k=\gr(\psi_{k-1})$, we get from Equation~\eqref{eq:sql-update-2}
\begin{equation}
    \psi_k = \psi_{k-1} + T_{\pi_k} \psi_{k-1} - \gamma P_{\pi_{k-1}} \psi_{k-2} + \epsilon_k,
\end{equation}
which is exactly Equation~\eqref{eq:psi-sql}.

\section{Experiment details on Garnets \label{app:garnets}}

\subsection{Sampled MoVI}
We provide the pseudo-code for the algorithm used in Section~\ref{sec:sample} in Algotithm~\ref{algo:movi}.
\begin{algorithm}
\caption{Sampled MoVI}
\begin{algorithmic}[1]
\label{algo:movi}
\REQUIRE K number of iterations. Initialize $q_0 = 0^{\states \times \actions}$, $h_0 = q_0$, $\pi_1 \in \gr(h_0)$
\FOR{$k = 1$ \TO $K$}
    \FOR{$(s,a) \in \states\times\actions$}
       \STATE $\p{s} \sim P(\cdot|s,a)$
       \STATE $q_k(s,a) = r(s,a) + \gamma q_{k-1}(\p{s},\pi_k(\p{s}))$
    \ENDFOR
    \STATE $h_k = \frac{k}{k+1}h_{k-1} + \frac{1}{k+1}q_k$
    \FOR{$s \in \states$}
        \STATE $\pi_{k+1}(s) = \argmax(h_{k}(s,\cdot))$    
    \ENDFOR
\ENDFOR
\RETURN $\pi_{K}$
\end{algorithmic}
\end{algorithm}

\subsection{Experiment details on Garnets}
For our experiments, we averaged the results over $100$ Garnets built with  $N_S =30$ (number of states), $N_a = 4$ (number of actions), and $N_B=4$ (branching factor).

\paragraph{Assumption check.} We provide in Fig.~\ref{fig:std-adp} the graphs showing the standard deviation of $\hat{\epsilon}_N$ over $100$ garnets. We also put the empirical means to be clearer.
\begin{figure}
\begin{minipage}{.5\linewidth}
    \centering
    \includegraphics[width=\linewidth]{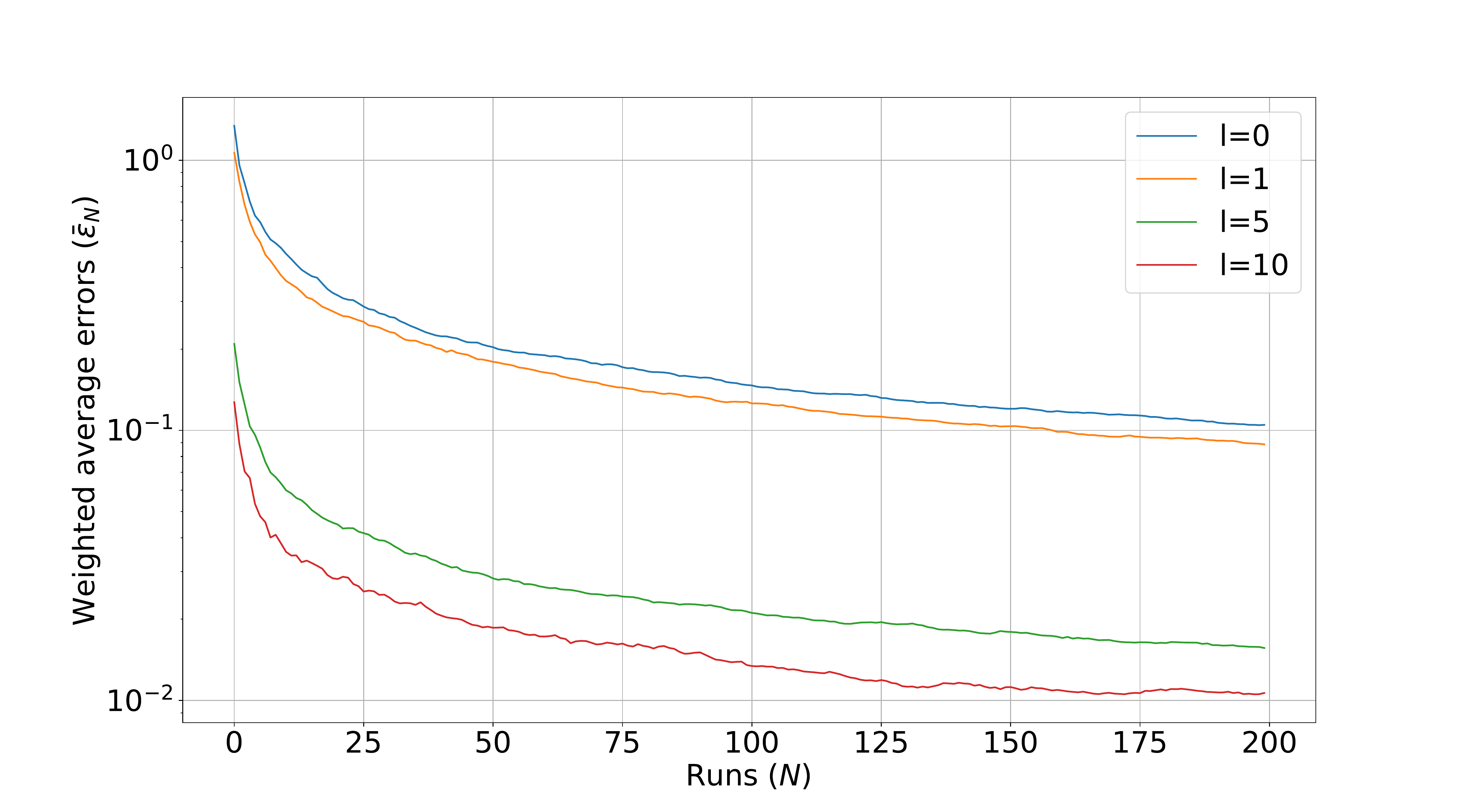}
\end{minipage}
\begin{minipage}{.5\linewidth}
    \centering
    \includegraphics[width=\linewidth]{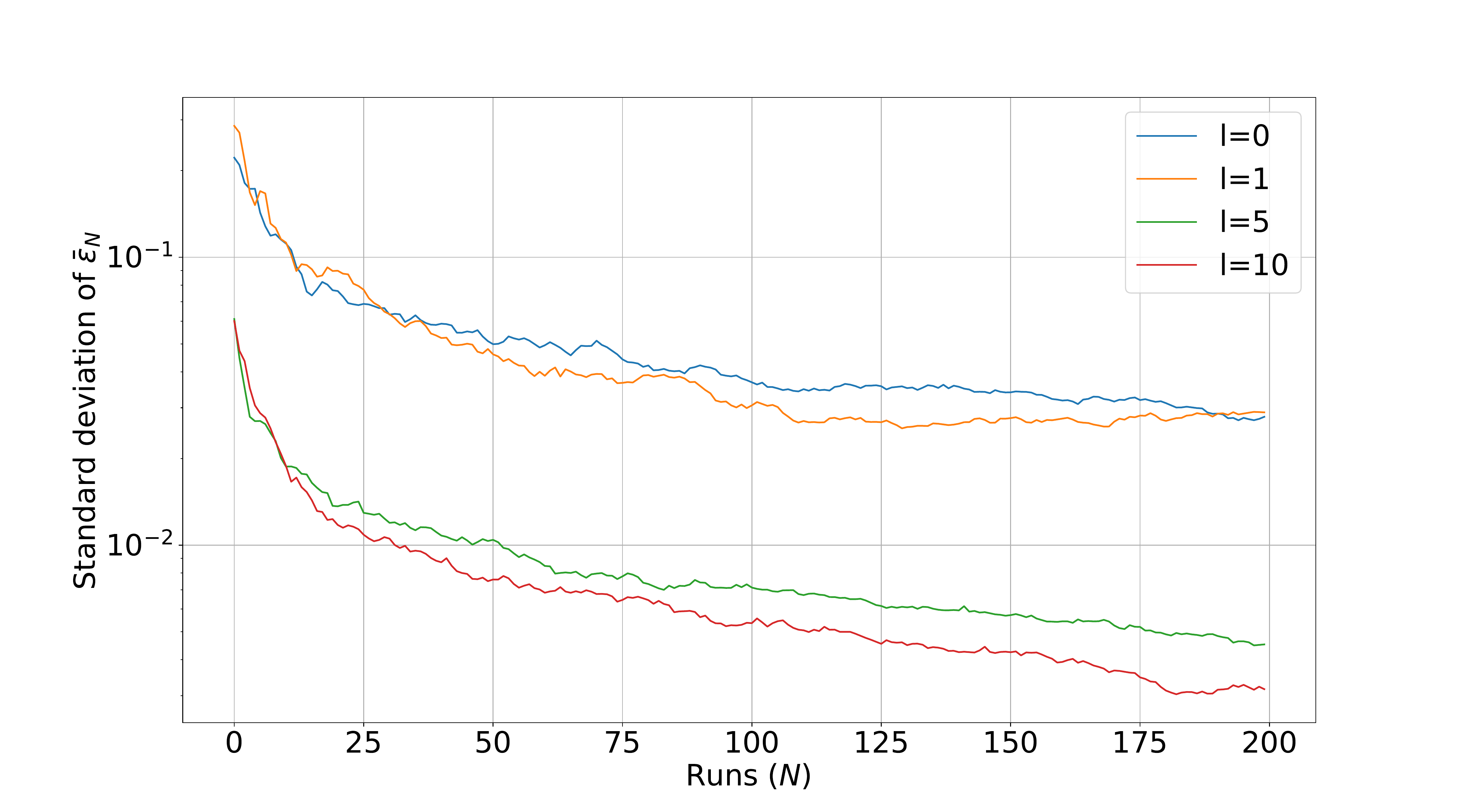}
\end{minipage}
\caption{\textbf{Left:} Empirical mean of $\hat{\epsilon}_{l,N}$ over garnets. \textbf{Right:} Standard deviation  over garnets for different values of $l$. Results are computed over on $100$ Garnets.} 
\label{fig:std-adp}
\end{figure}

\paragraph{Algorimths comparison.}In addition, we provide in Figure~\ref{fig:alg2-adp} the graphs showing the standard deviation over MDPs of the curves computed from Figure~\ref{fig:adp}, where we compared the different ADP schemes (we also put the means again, to be clearer).
\begin{figure}
\begin{minipage}{.5\linewidth}
    \centering
    \includegraphics[width=\linewidth]{figures/momentum_dpp_sql_vi_10000_pol.pdf}
\end{minipage}
\begin{minipage}{.5\linewidth}
    \centering
    \includegraphics[width=\linewidth]{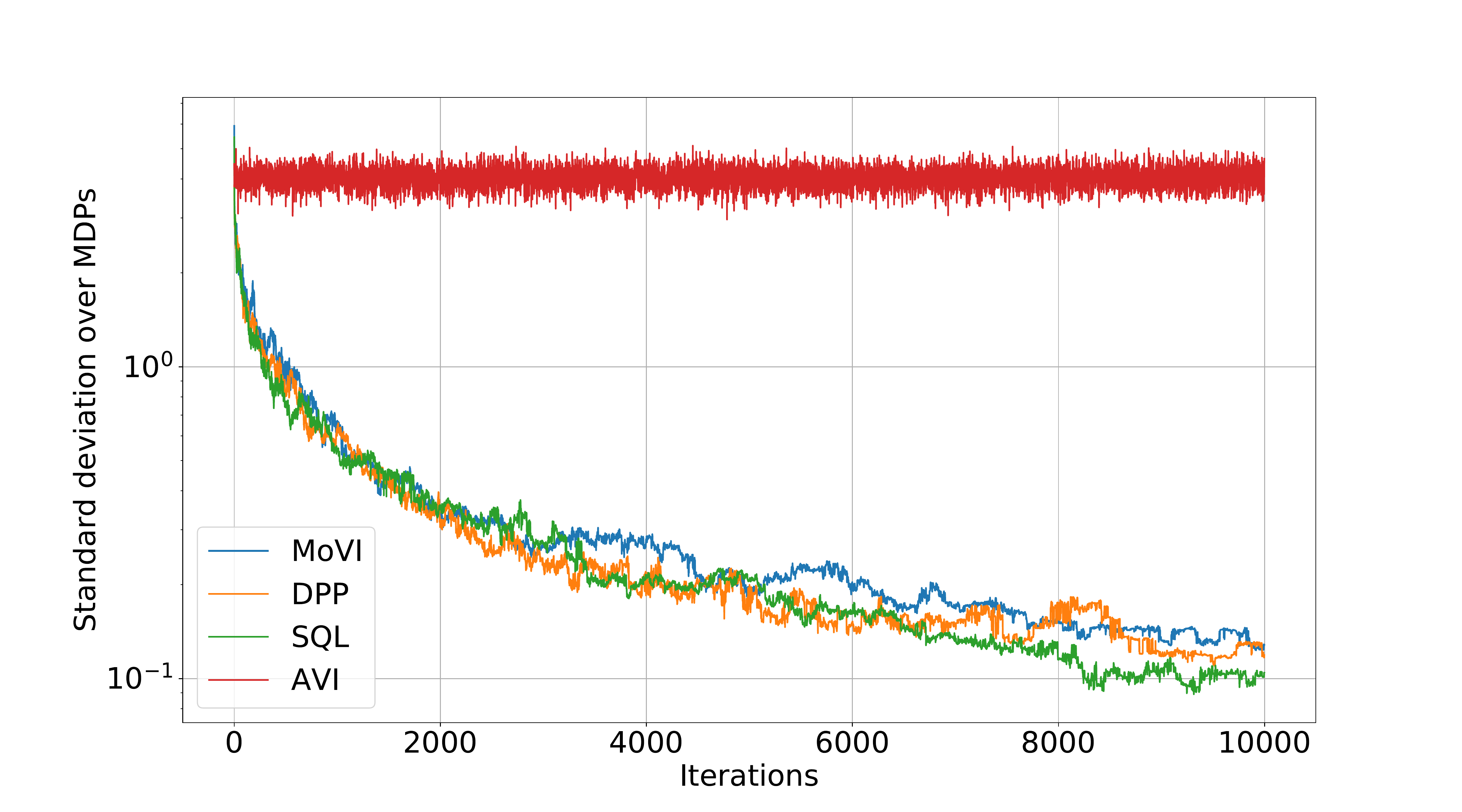}
\end{minipage}
\caption{\textbf{Left:} Mean of the average errors over garnets. \textbf{Right:} Standard deviation of the average errors over garnets of AVI, MoVI, DPP and DSQL. Resulst are computed over on $100$ Garnets.} 
\label{fig:alg2-adp}
\end{figure}

\section{Additional Experiments on Atari \label{app:atari}}
In this section, we provide additional experiments on Atari, and details about experiments presented in the paper.

\subsection{DSQL and DDPP \label{app:atari-dsql}}
We implemented deep versions of SQL and DPP. we use the same parametrization as for Momentum-DQN (Section~\ref{sec:modqn}), without needing an extra $H$-network.

\paragraph{DSQL.} For DSQL, we keep two target networks, the copies of the two previous weight updates, repectively $Q^{-1} = \old{Q}$ and $Q^{-2} = (\old Q)^-$. We define a regression target for the $Q$-network as 
\begin{equation}
    \hat{Q}_{\text{dsql}}(s,a,r,\p{s}) = Q^{-1}(s,a) + (1 - \beta_k)(r + \gamma \max_{\p{a}}Q^{-2}(\p{s}, \p{a}) - Q^{-1}(s,a)) + \beta_k\gamma(\max_{\p{a}}Q^{-2}(\p{s}, \p{a}) - \max_{\p{a}}Q^{-1}(\p{s},\p{a})),
\end{equation}
which leads to a loss function on the weights $\theta$,
\begin{equation}
    \mathcal{L}_{\text{dsql}}(\theta) = \avg_\mathcal{B}\left[(\hat{Q}_{\text{dsql}}(s,a,r,\p{s}) - Q(s,a))^2\right].
\end{equation}

\paragraph{DDPP.} We proceed similarly for DDPP, keeping the same parametrization of the $Q$-network. We define a loss function on the weights
\begin{equation}
    \mathcal{L}_{\text{ddpp}}(\theta) = \avg_\mathcal{B}\left[(\old{Q}(s,a) + r + \gamma \max_{\p{a}} \old{Q}(\p{s},\p{a}) - \max_{\p{a}}\old{Q}(s,\p{a}) - Q(s,a))^2\right].
\end{equation}

\subsection{Experiment details \label{app:atari-exp}}

In Table~\ref{tab:atari}, we give the hyperparameters used for our experiments on Atari, including networks architecture. We use the following notations to describe neural networks: $\FC n$ is a fully connected layer with $n$ neurons; $\Conv_{a,b}^{d}c$ is a 2d convolutional layer with $c$ filters of size $a  \times b$ and a stride of~$d$. $n_A$ is the number of actions available in a game. We highlight the fact that we used the standard Dopamine's DQN parameters, and did not try to optimize them (including the optimizer of the additional neural network).

\begin{table}%
    \centering
    \caption{Parameters used for Momentum-DQN on Atari. the $Q$-network and $H$-network have the same structure.}
    \begin{tabular}{ll}
    \hline
    Parameter     & Value \\
    \hline
    $C$ (update period)    & 8000\\
    $F$ (interaction  period)    & 4\\
    $\gamma$ (discount) & 0.99\\
    $|\mathcal{B}|$ (replay buffer size) & $10^6$\\
    $|B_{h,k}|$ and $|B_{q,k}|$ (batch size) & 32 \\
    $e_k$ (random actions rate) & 0.01 (with a linear decay of period $2.5\cdot10^5$ steps)\\
    $\kappa$ (mixture rate update period) & $2.5 \cdot 10^6$\\
    $Q$-network structure & $\Conv_{8,8}^{4}32-\Conv_{4,4}^{2}64-\Conv_{3,3}^{1}64-\FC512-\FC n_A$\\
    activations & Relu\\
    optimizers & RMSprop ($lr=0.00025$) \\
    \hline
    \end{tabular}
    \label{tab:atari}
\end{table}

\subsection{Influence of the mixture rate \label{app:atari-mixt}} 
We look at the influence of the sequence $\beta_k$ on Momentum-DQN. To do that, we first test Momentum-DQN($\beta$), a version of Algorithm~\ref{algo:mom-dqn} with $\beta_k = \beta$ for each $k$. We show training curves of this algorithm in in Figure~\ref{fig:beta}, where we evaluate Momentum-DQN($\beta$) on the games Asterix and Zaxxon with different values of $\beta$. We report results for $\beta=0.1$, $0.5$, and $0.9$. We observe for Asterix how a higher $\beta$ -- meaning more influence of the old $Q$-networks -- slows down learning in the beginning, but eventually leads to a much higher performance. We also observed that a very high $\beta$ (close to $1$) tends to slow training so much it affects drastically the sample complexity of Momentum-DQN. On some games, like Zaxxon in Fig.~\ref{fig:beta}, it also seemed that Momentum-DQN needed a more aggressive update in the beginning, meaning a lower $\beta$. During this experiments, we observed  in general that there was an optimal $\beta$ per game, that could be quite different from game to game, so we could have higher results with a problem-dependent parametrization. For example, in Figure~\ref{fig:beta}, we observe how a high $\beta$ helps DQN on Asterix, but can damage Zaxxon if too close to $1$.

This observation justifies the utilization of a schedule of increasing $\beta_k$ as described in Section~\ref{sec:modqn}. The schedule defined here is inspired by the the theory, but we could also imagine a heuristic increasing schedule, with for example $\beta_k = 0.1*\floor{\frac{k}{\kappa}}$. We tested similar schedules that gave almost the same performance as the one presented here.

\begin{figure}
\begin{minipage}{.5\linewidth}
    \centering
    \includegraphics[width=\linewidth]{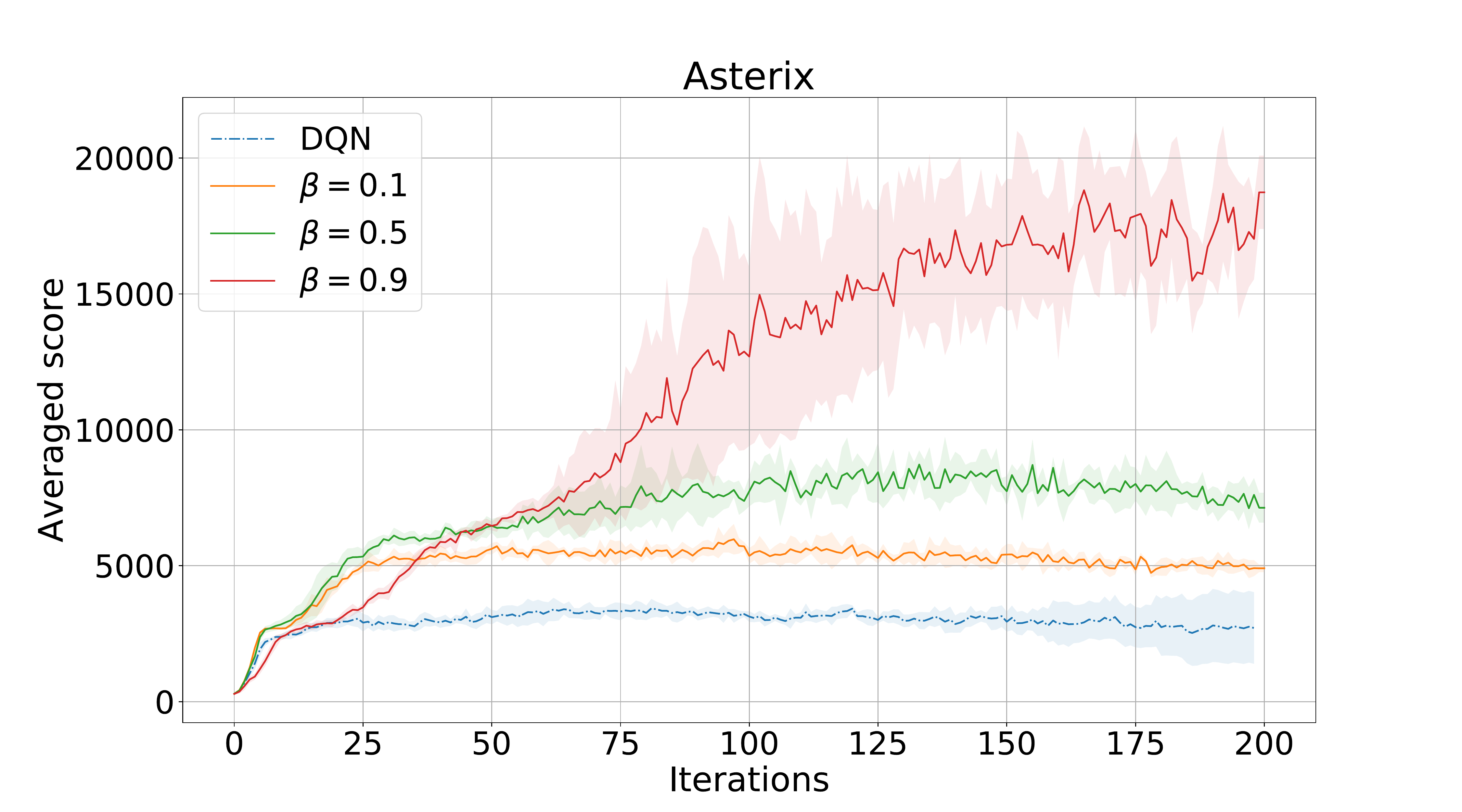}
\end{minipage}
\begin{minipage}{.5\linewidth}
    \centering
    \includegraphics[width=\linewidth]{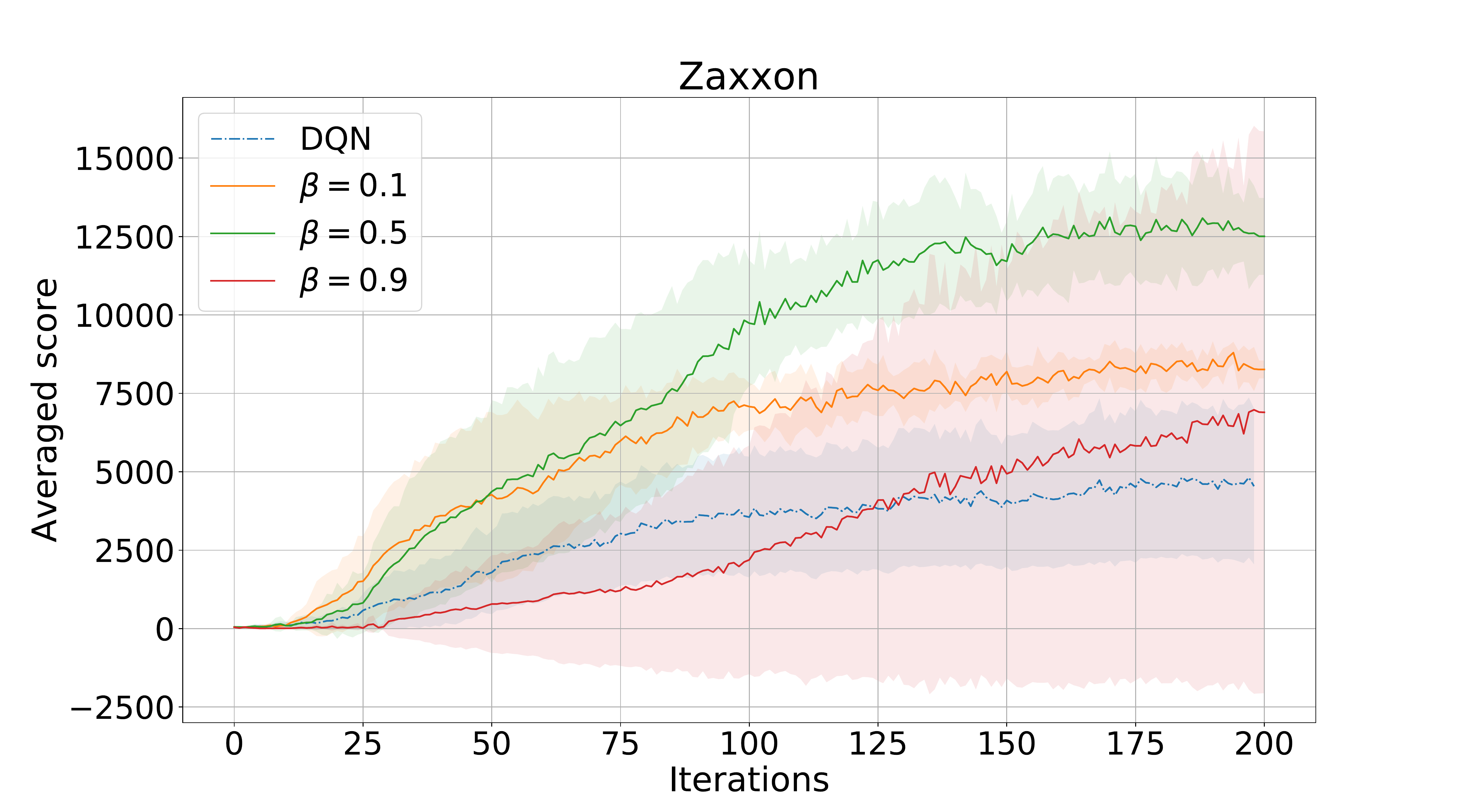}
\end{minipage}
\caption{Influence of $\beta$ on Asterix (\textbf{left}) and Zaxxon (\textbf{right}). Each curve is shows the evolution of the score obtained by Momentum-DQN trained with a fixed $\beta$. In blue and dashed-dot line, we show the DQN baseline. In orange $\beta=0.1$, in green $\beta=0.5$, in red $\beta=0.9$. This shows how a higher $\beta$ slows learning in the beginning but allows for a higher final performance.} 
\label{fig:beta}
\end{figure}

\subsection{Joint comparison on Atari \label{app:atari-joint}}
We provide the joint graph showing the AUC improvement over DQN of Momentum-DQN, DSQl and DDPP in Figure~\ref{fig:joint}.
\begin{figure}
    \centering
    \includegraphics[width=.6\linewidth]{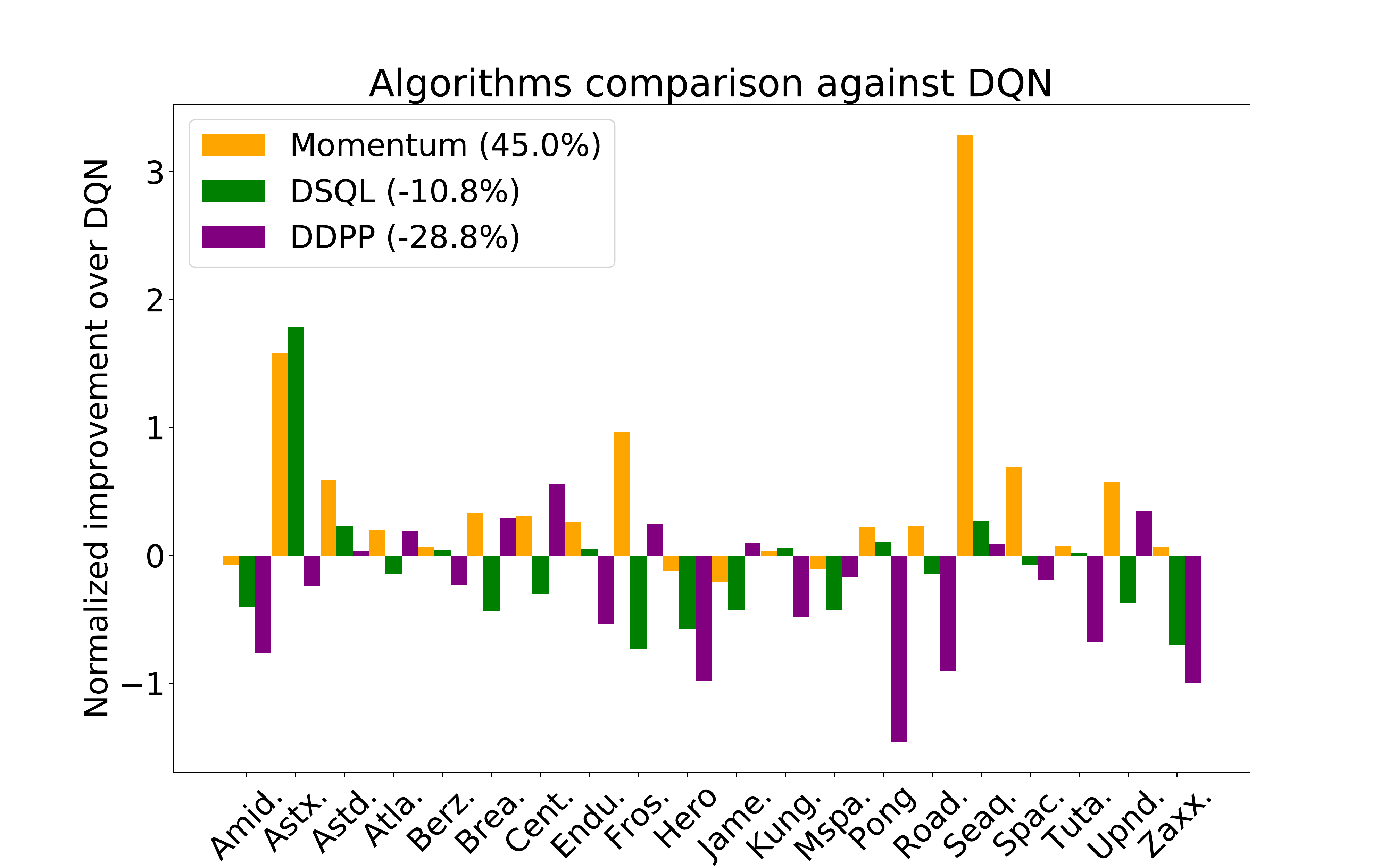}
    \caption{AUC improvement over DQN of Momentum-DQN (orange), DSQL (green) and DDPP (purple).} 
    \label{fig:joint}
\end{figure}

\subsection{Full results on Atari \label{app:atari-full}}
We provide the learning curves on the 20 considered Atari games, for Momentum-DQN, DSQL and DPP, compared with a DQN baseline, in Figures~\ref{fig:full} and~\ref{fig:full2}.

\begin{figure}
\begin{center}
\begin{tabular}{cc}
     \includegraphics[width=0.42\linewidth]{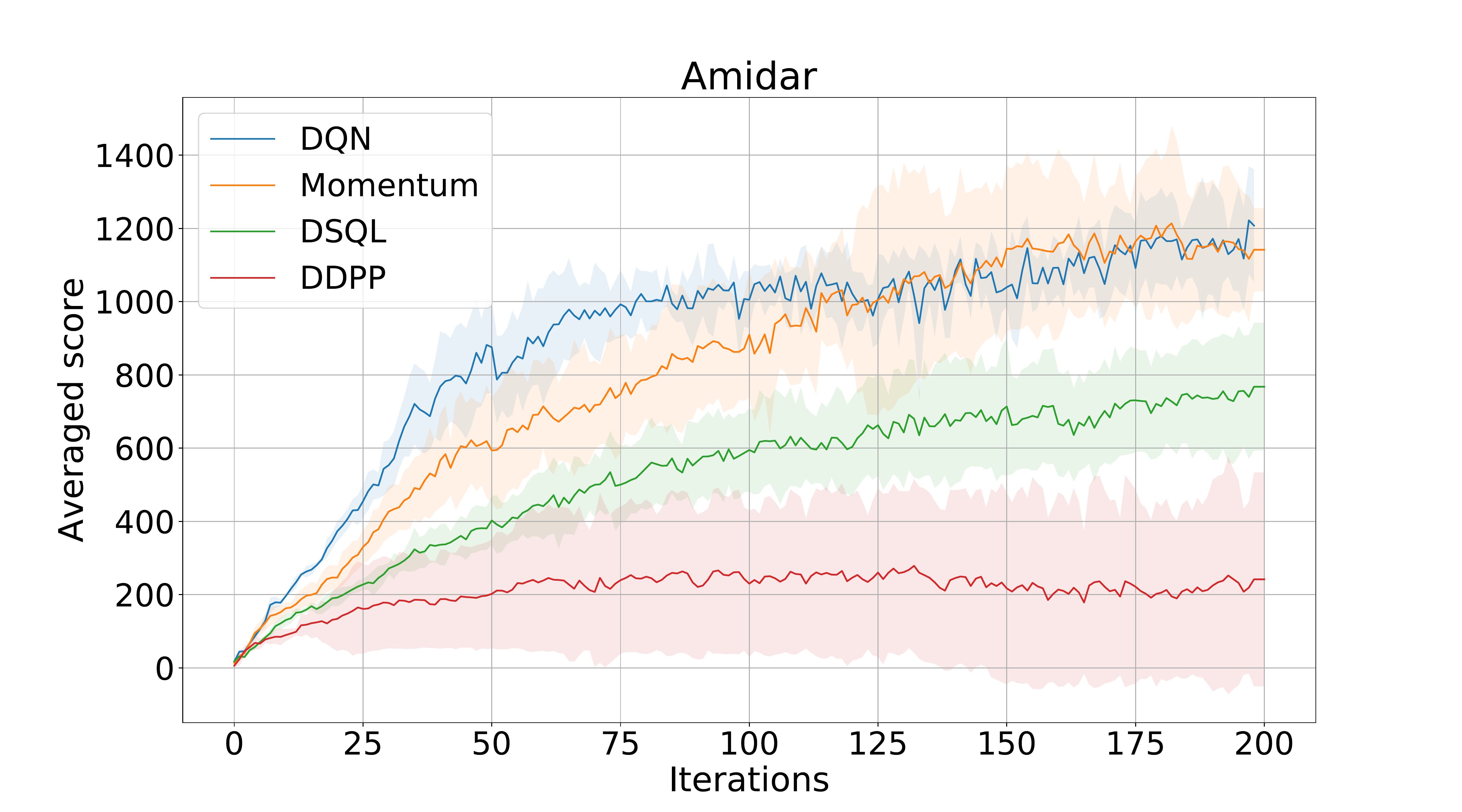} & \includegraphics[width=0.42\linewidth]{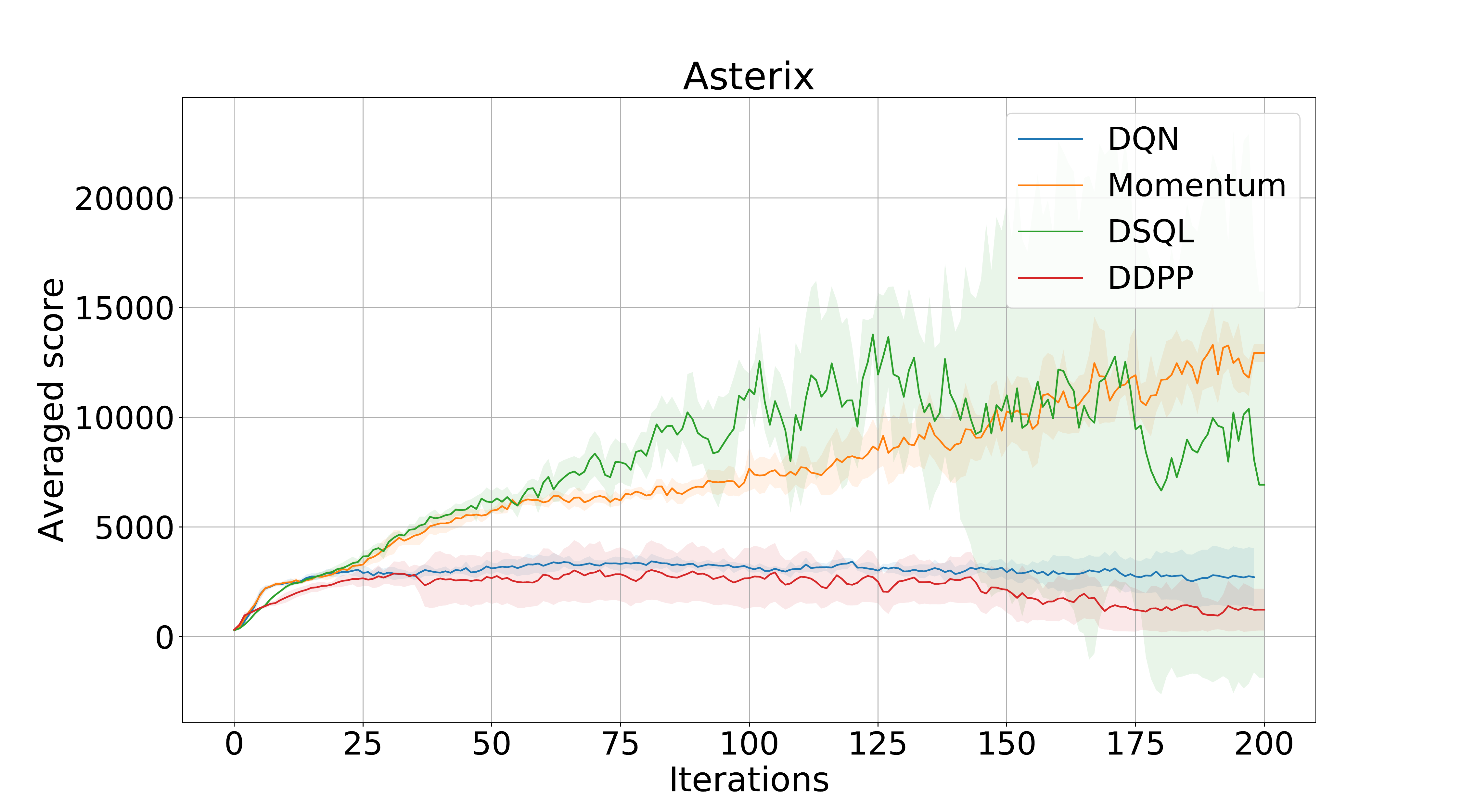}  \\
     Amidar & Asterix \\ 
     \includegraphics[width=0.42\linewidth]{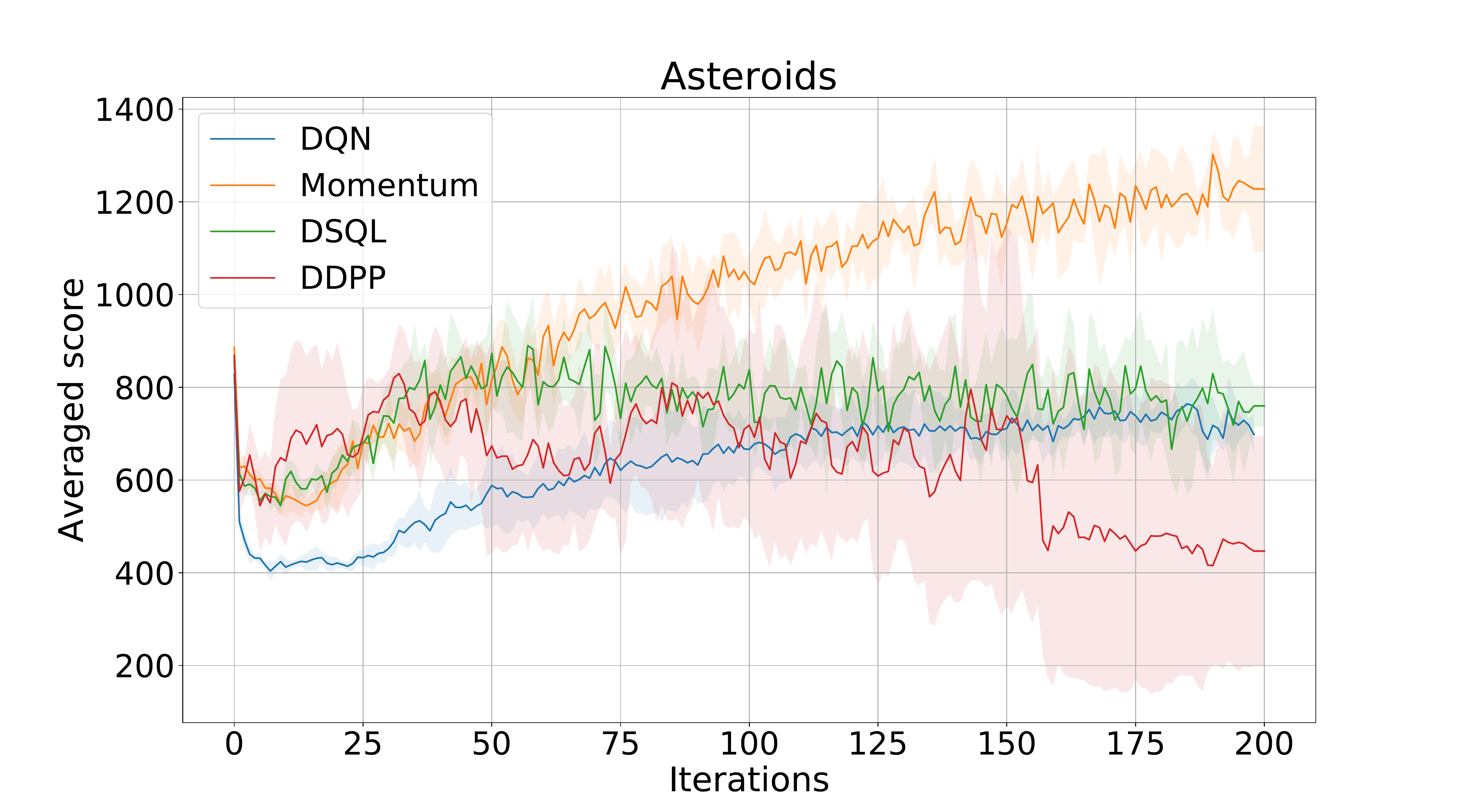} & \includegraphics[width=0.42\linewidth]{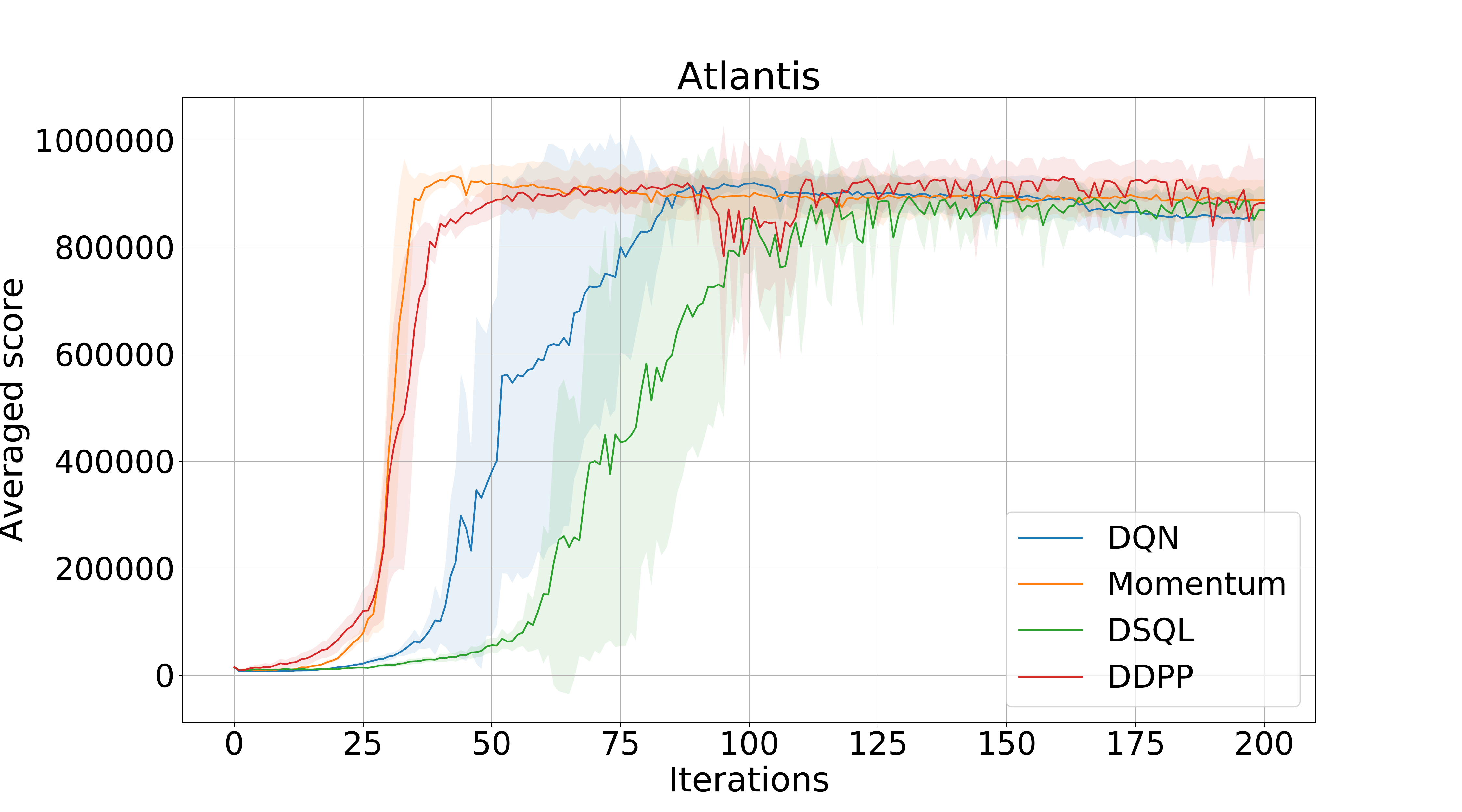}  \\
     Asteroids & Atlantis \\
     \includegraphics[width=0.42\linewidth]{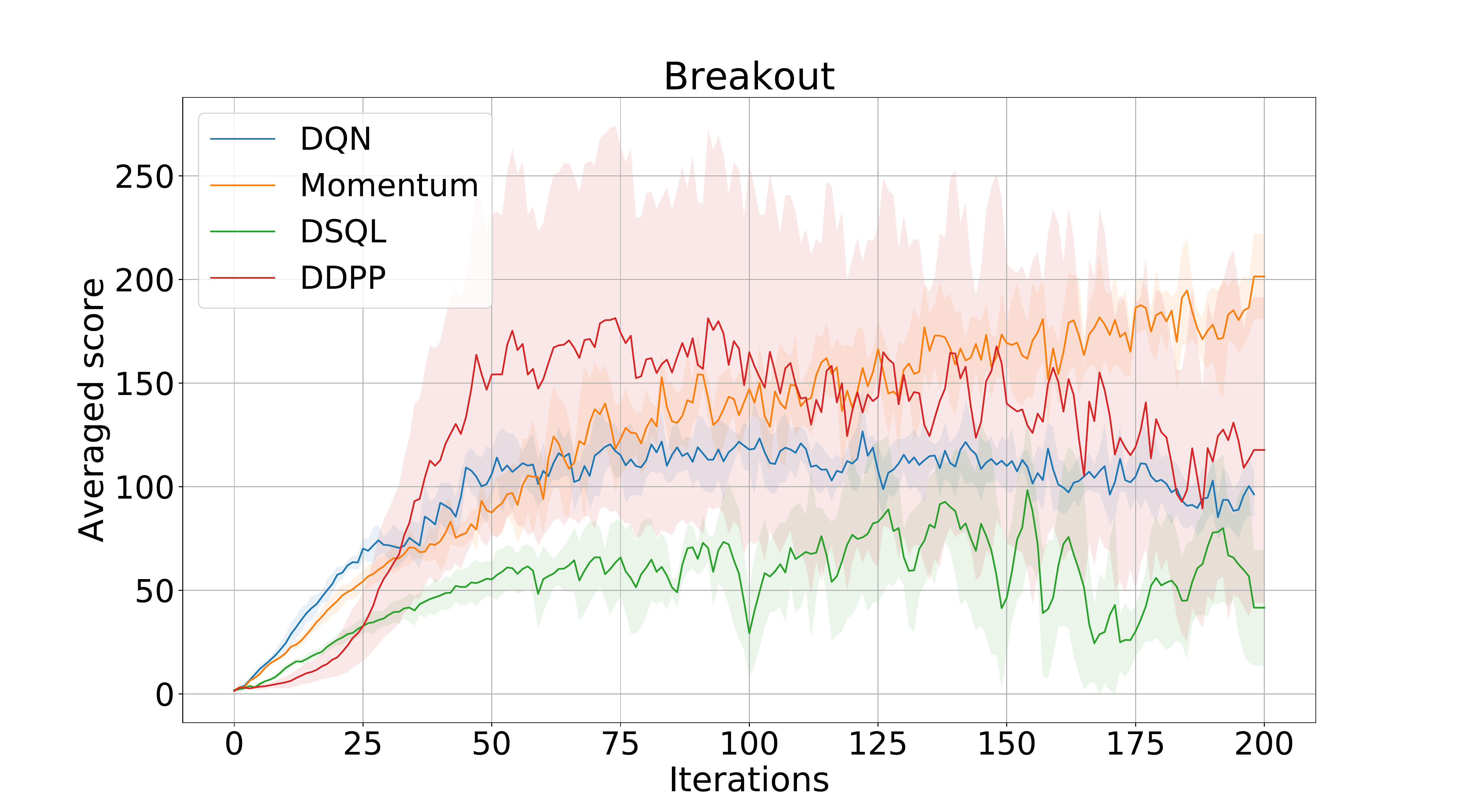} & \includegraphics[width=0.42\linewidth]{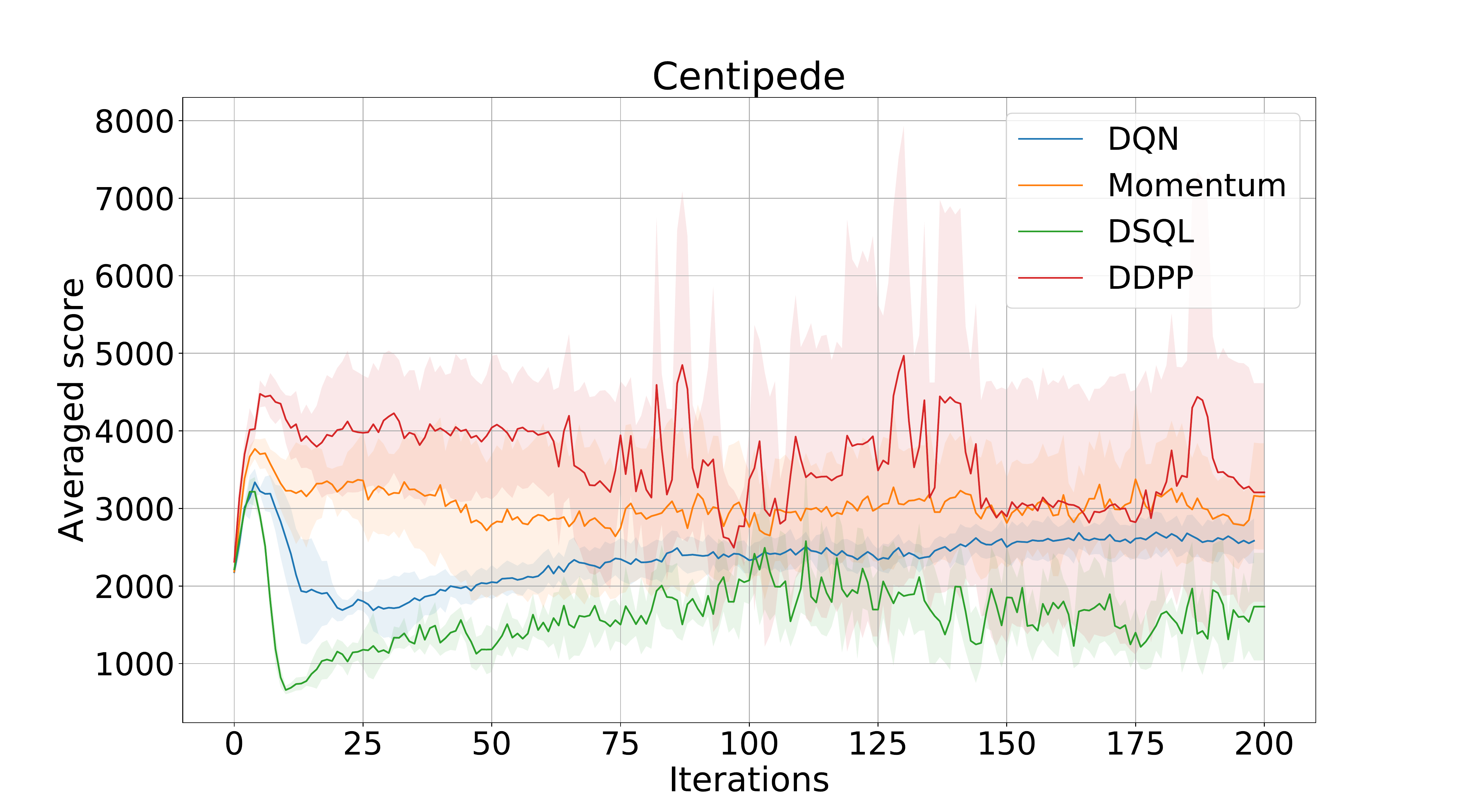}  \\
     Breakout & Centipede \\
     \includegraphics[width=0.42\linewidth]{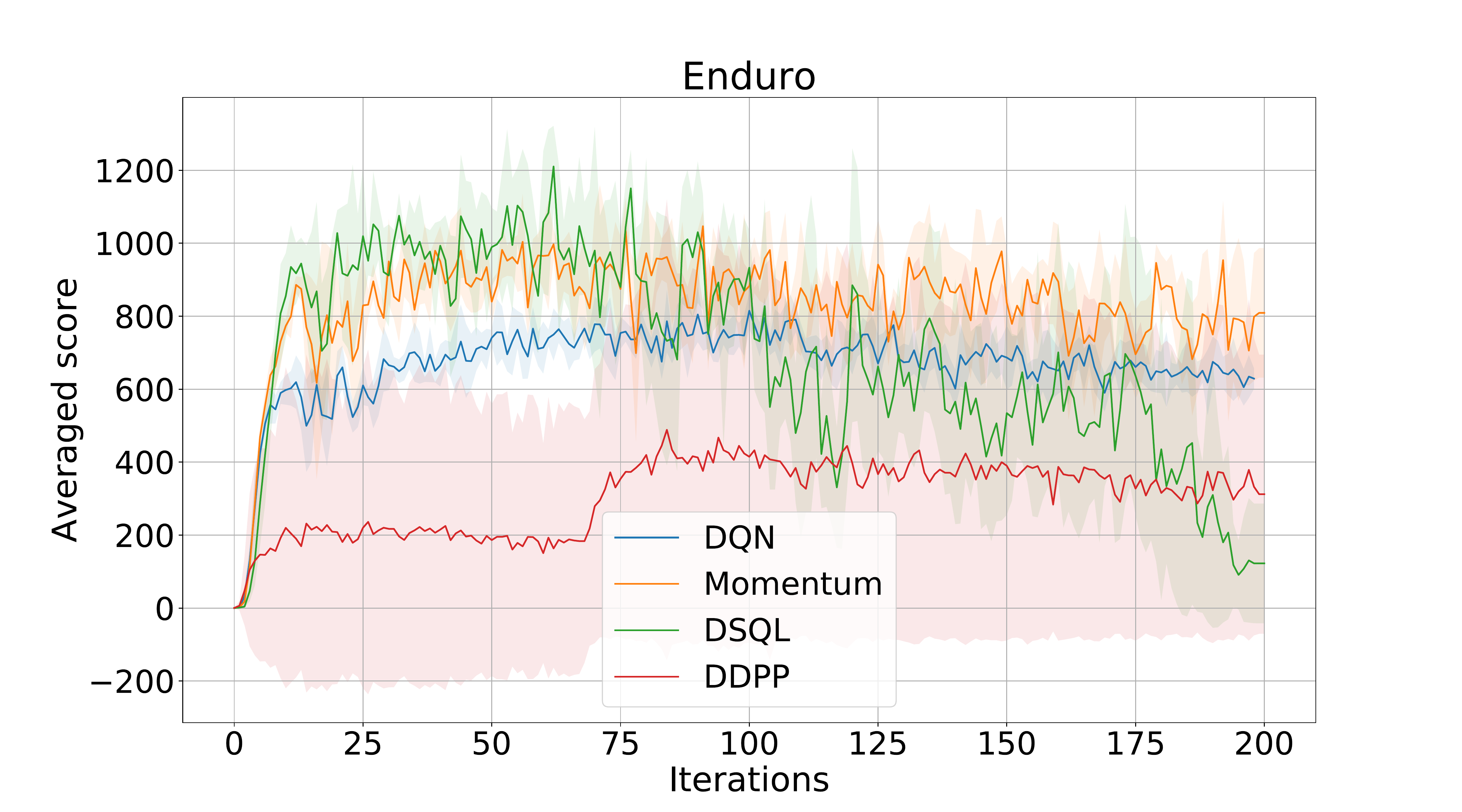} & \includegraphics[width=0.42\linewidth]{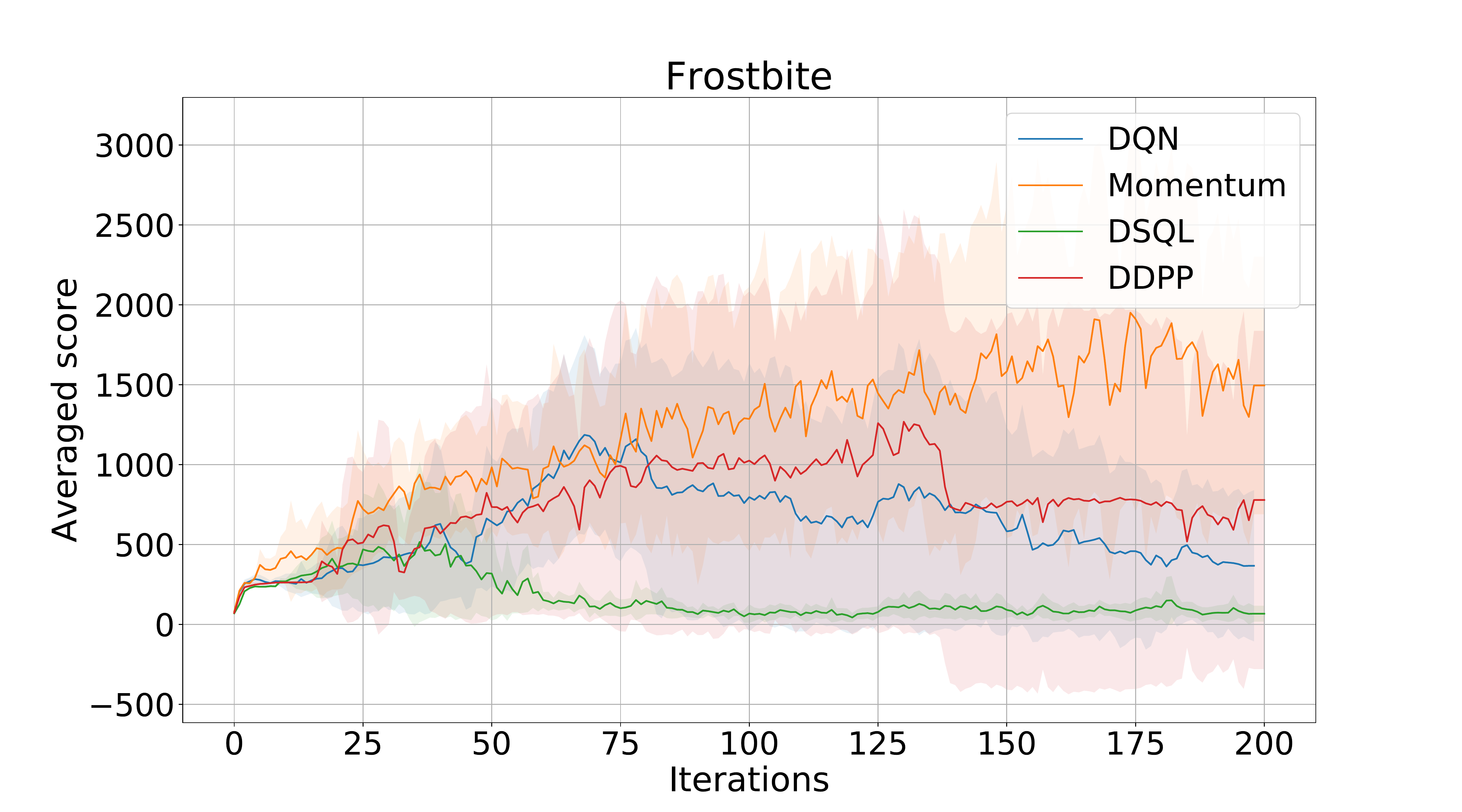}  \\
     Enduro & Frostbite \\
     \includegraphics[width=0.42\linewidth]{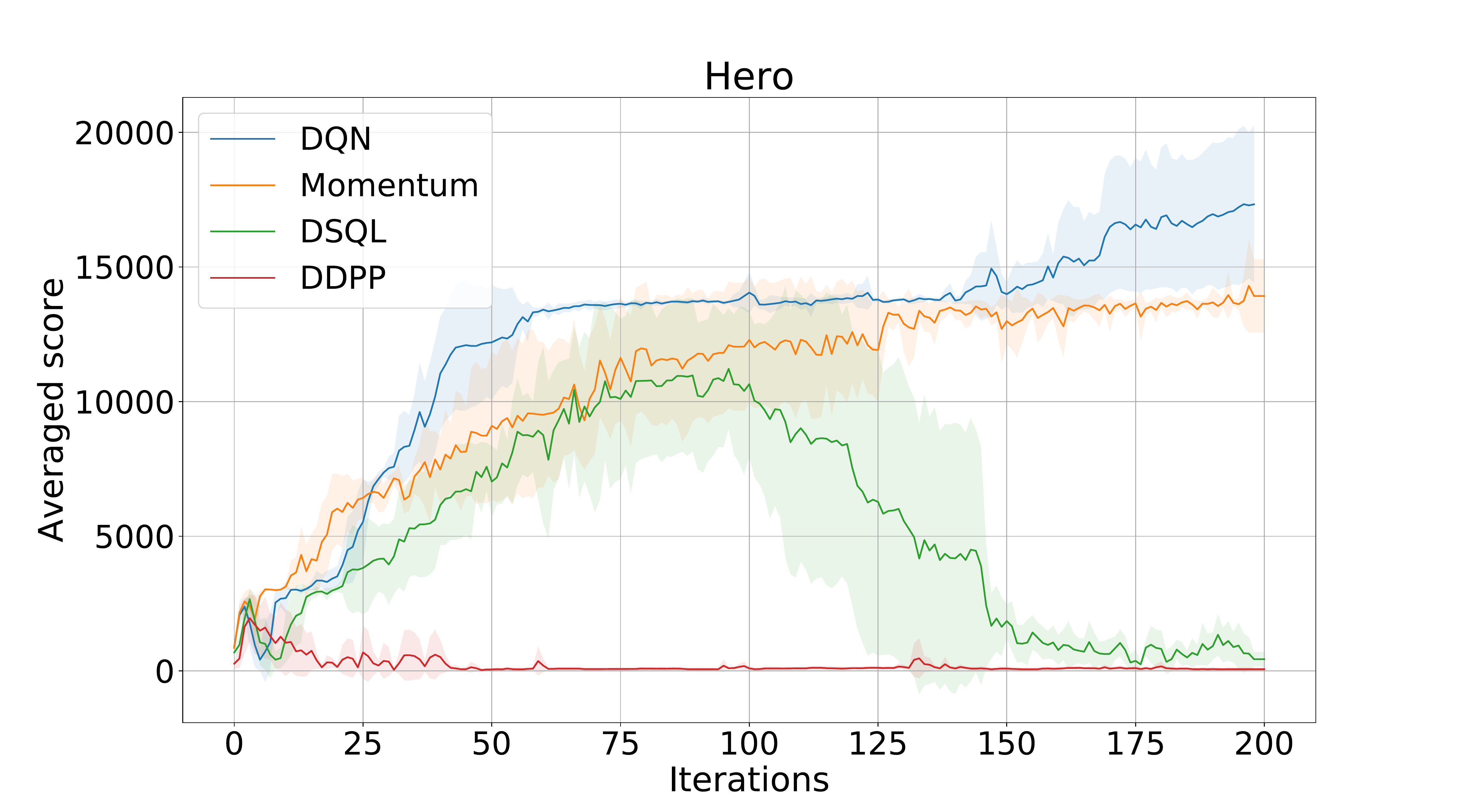} & \includegraphics[width=0.42\linewidth]{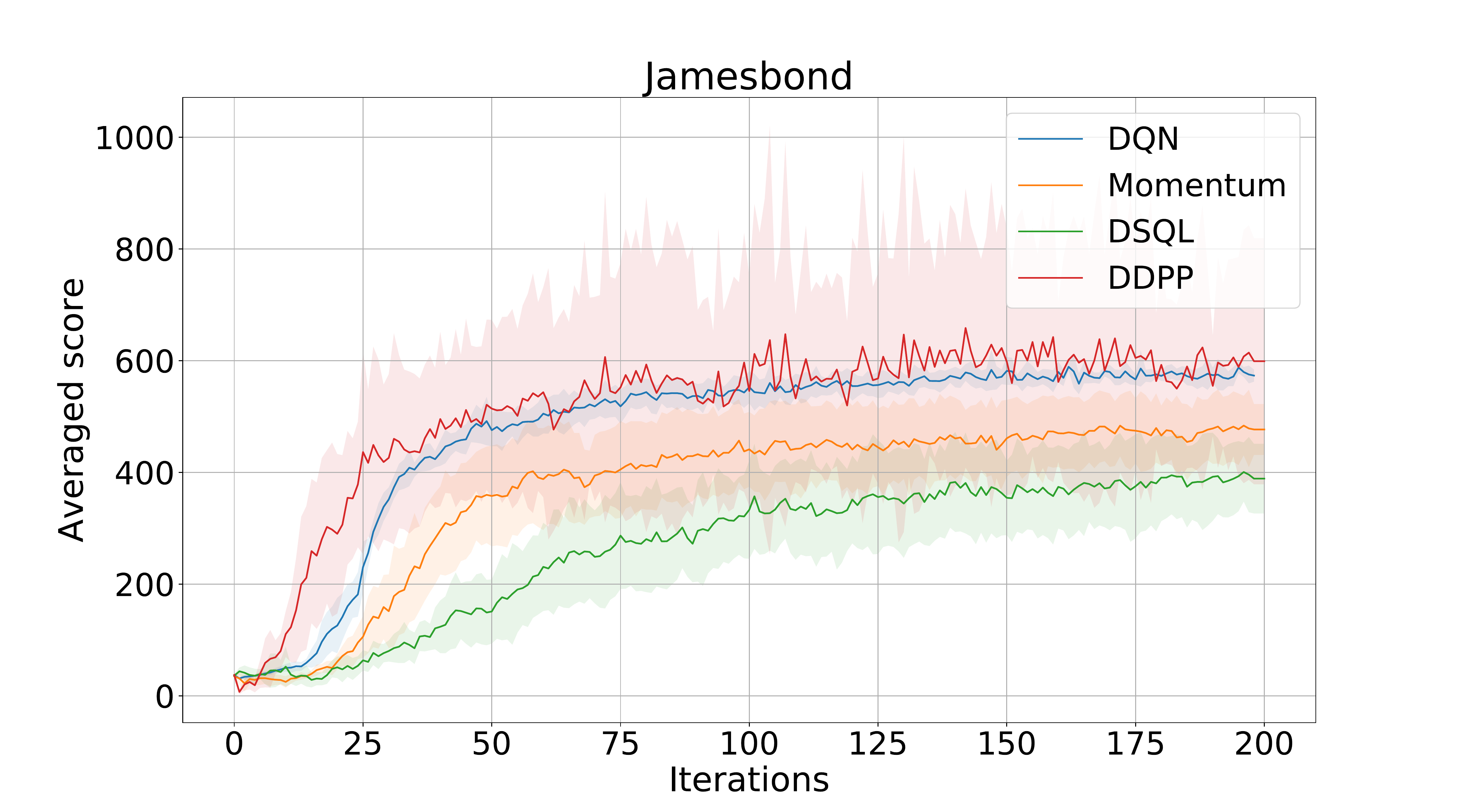}  \\
     Hero & Jamesbond \\
\end{tabular}
\caption{All averaged training scores of Momentum-DQN (orange), DSQL (green), DDPP (red), against DQN (blue)  on the subset of Atari games (1/2).\label{fig:full}}
\end{center}
\end{figure}

\begin{figure}
\begin{center}
\begin{tabular}{cc}
     \includegraphics[width=0.42\linewidth]{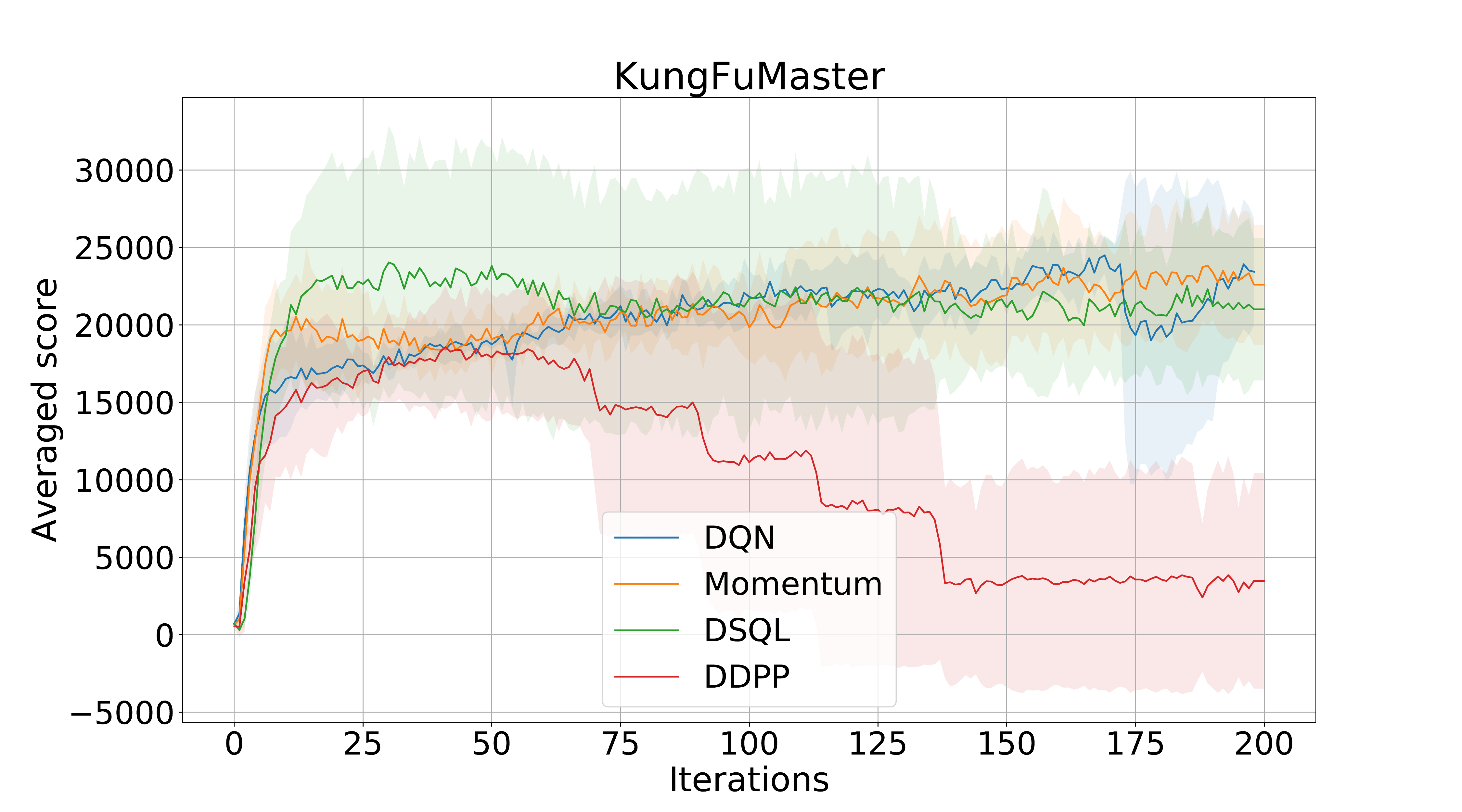} & \includegraphics[width=0.42\linewidth]{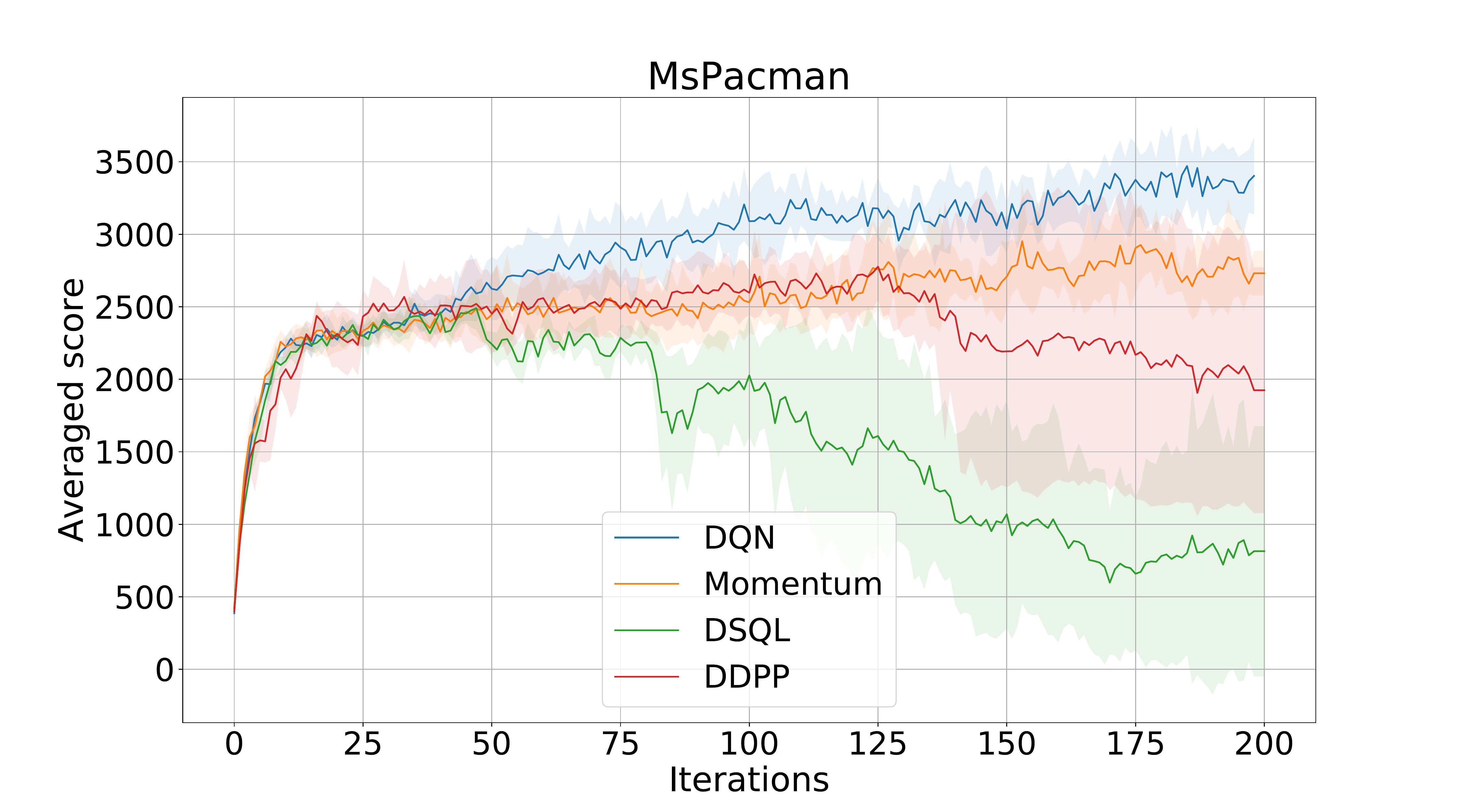}  \\
     KungFuMaster & MsPacman \\ 
     \includegraphics[width=0.42\linewidth]{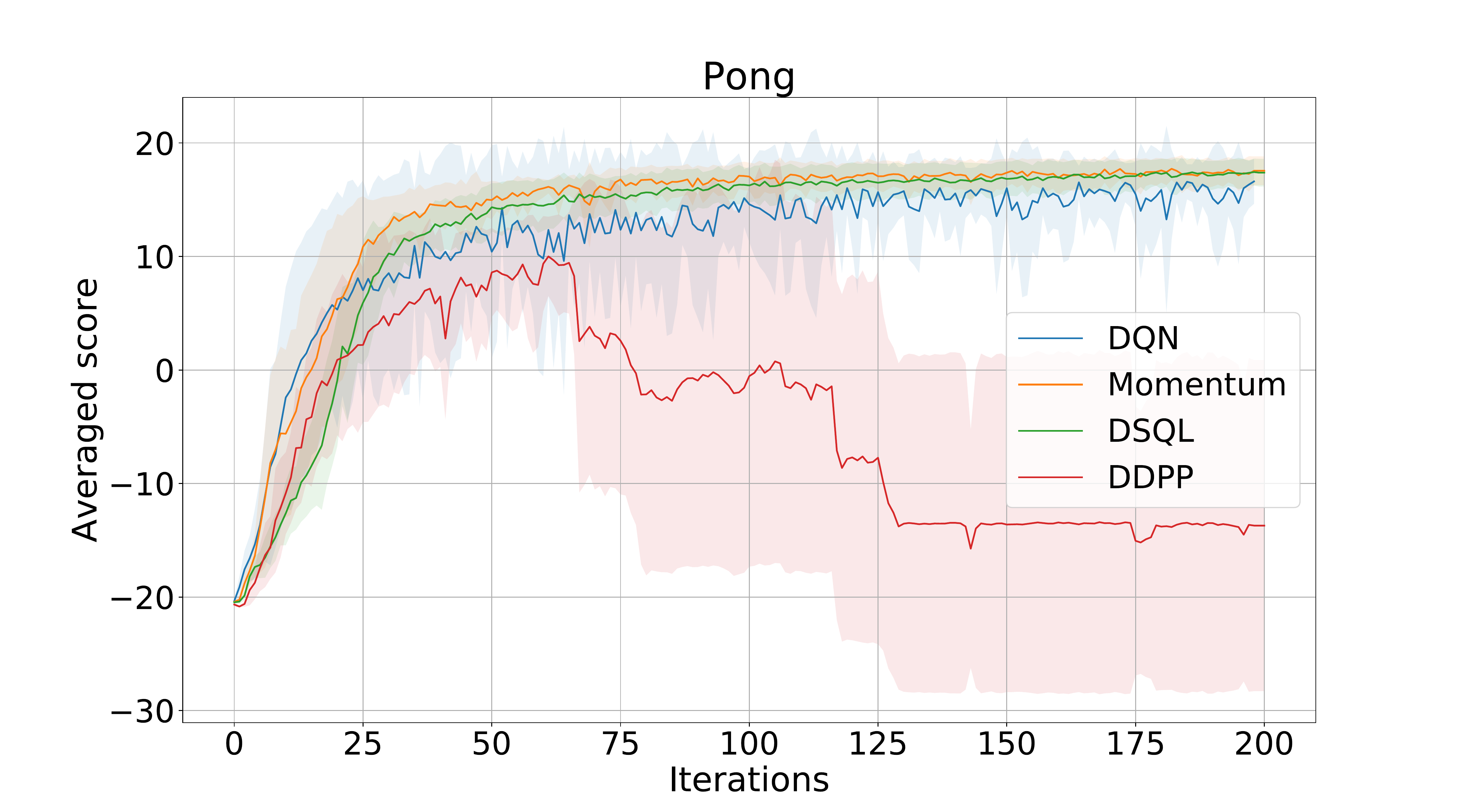} & \includegraphics[width=0.42\linewidth]{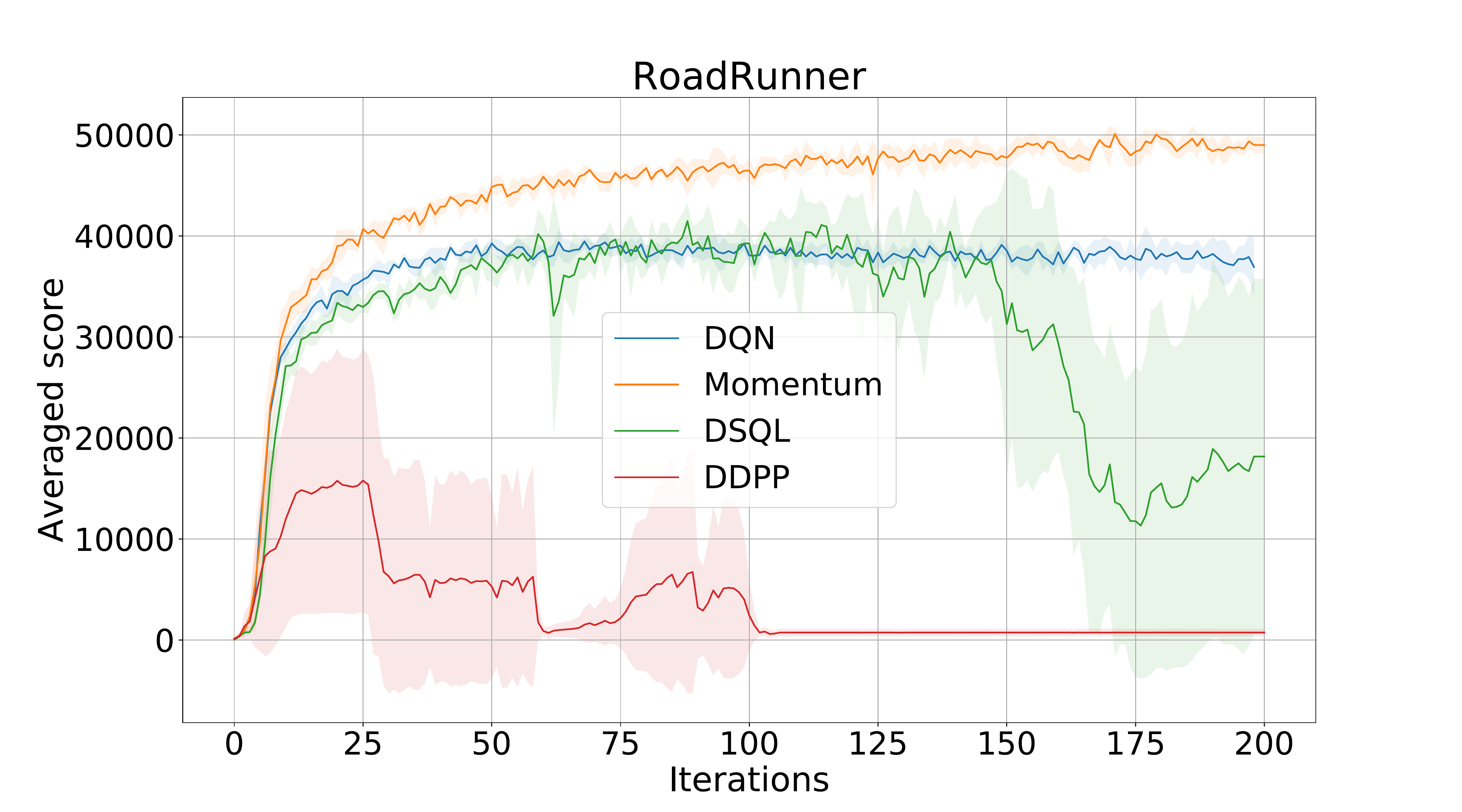}  \\
     Pong & RoadRunner \\
     \includegraphics[width=0.42\linewidth]{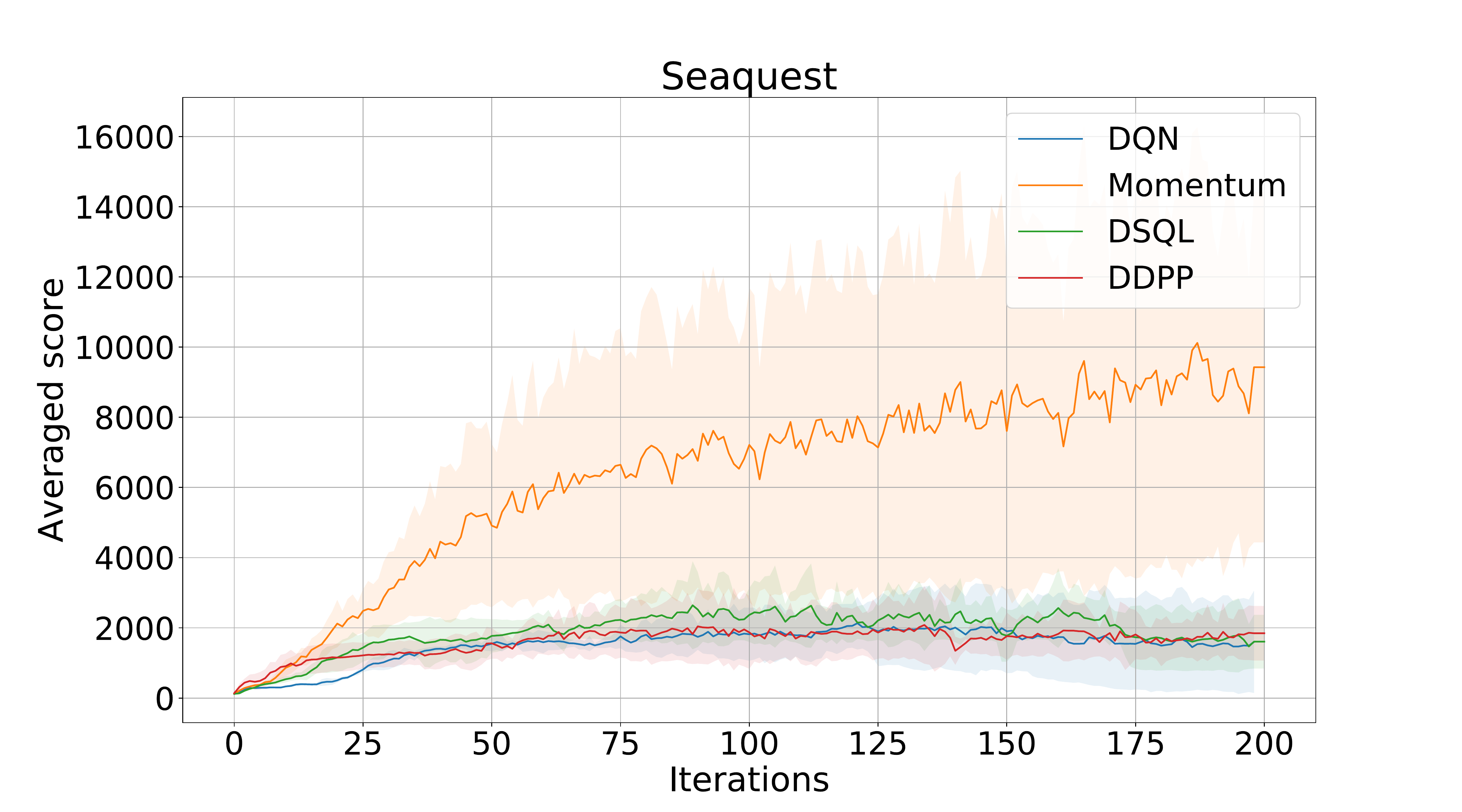} & \includegraphics[width=0.42\linewidth]{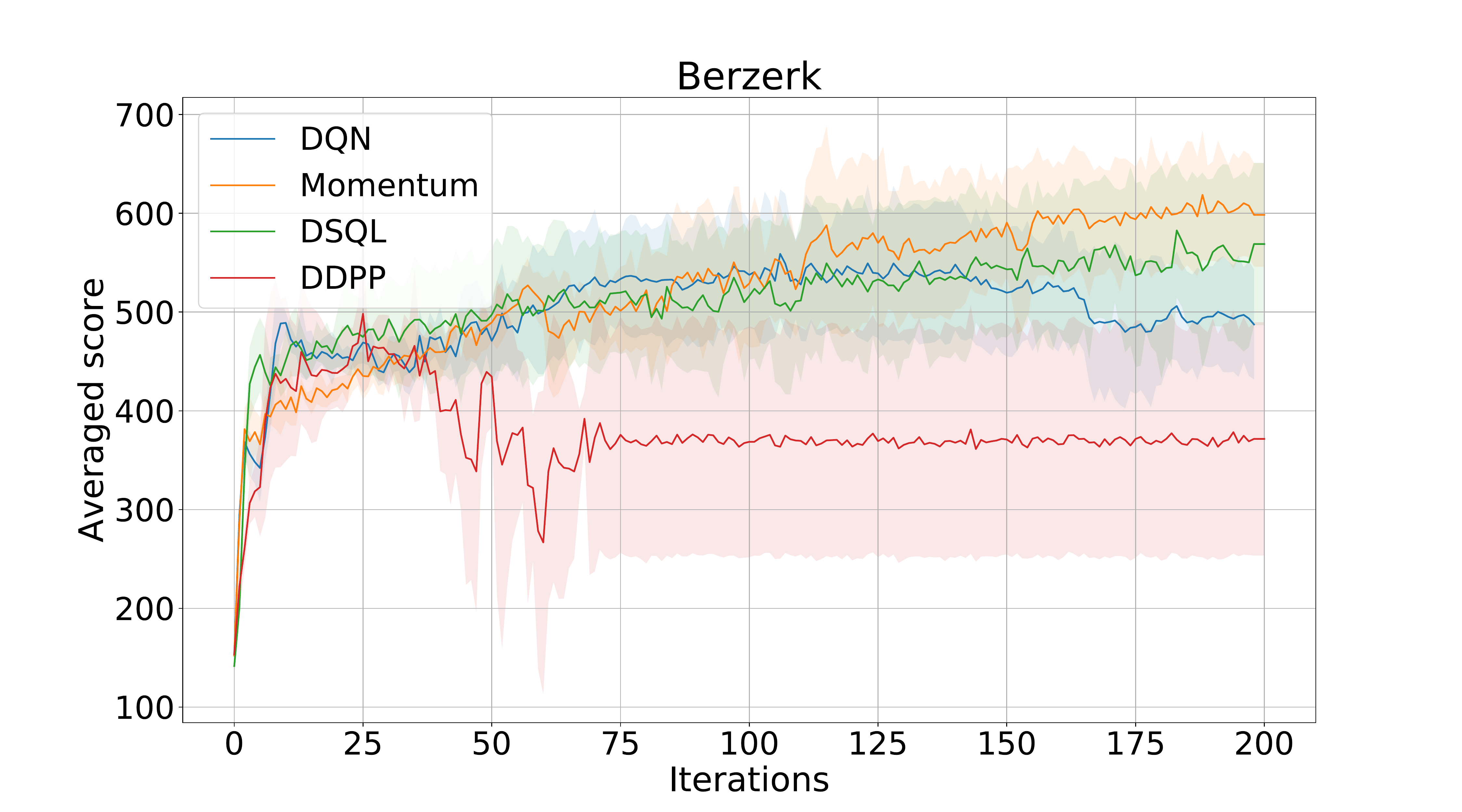}  \\
     Seaquest & Berzerk \\
     \includegraphics[width=0.42\linewidth]{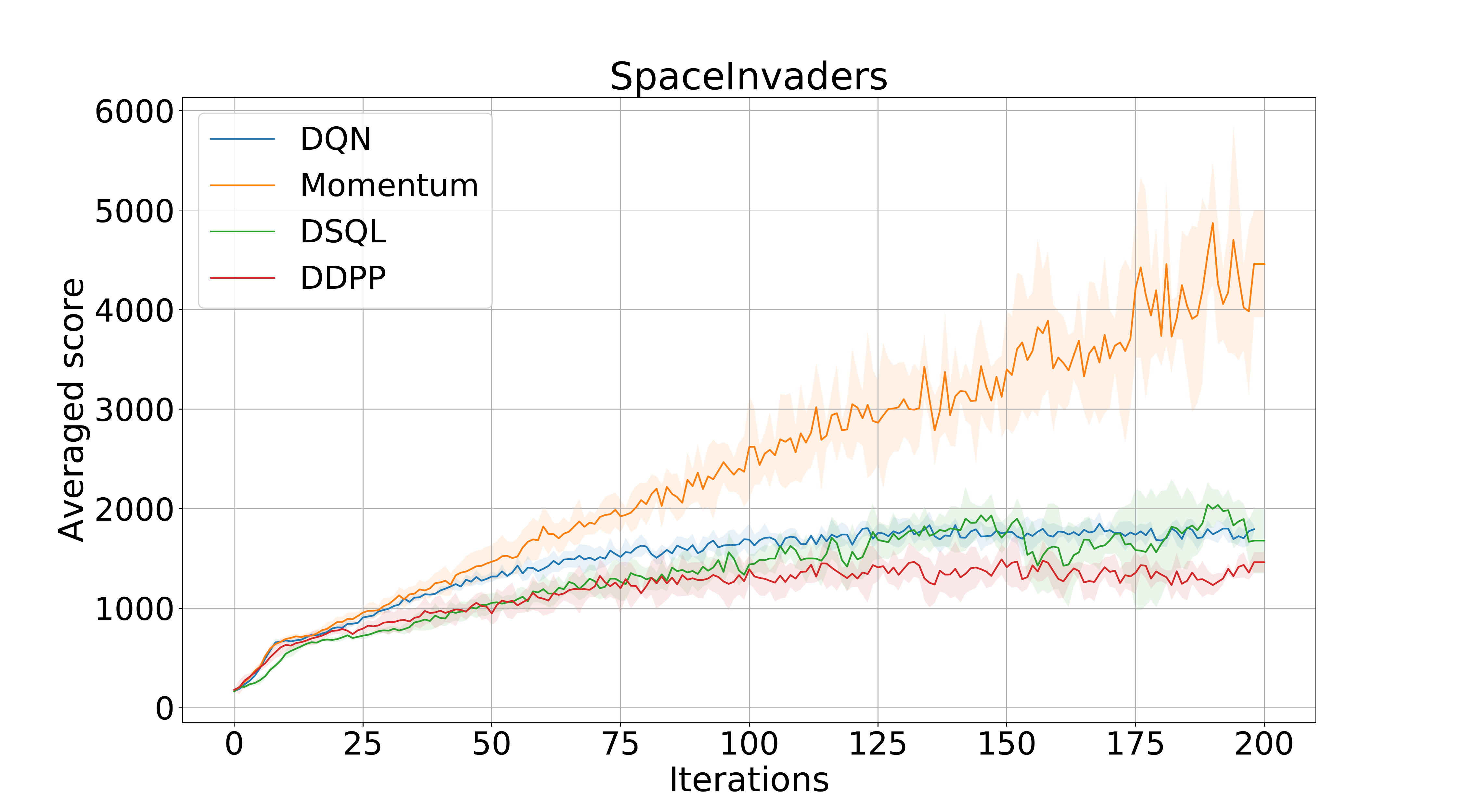} & \includegraphics[width=0.42\linewidth]{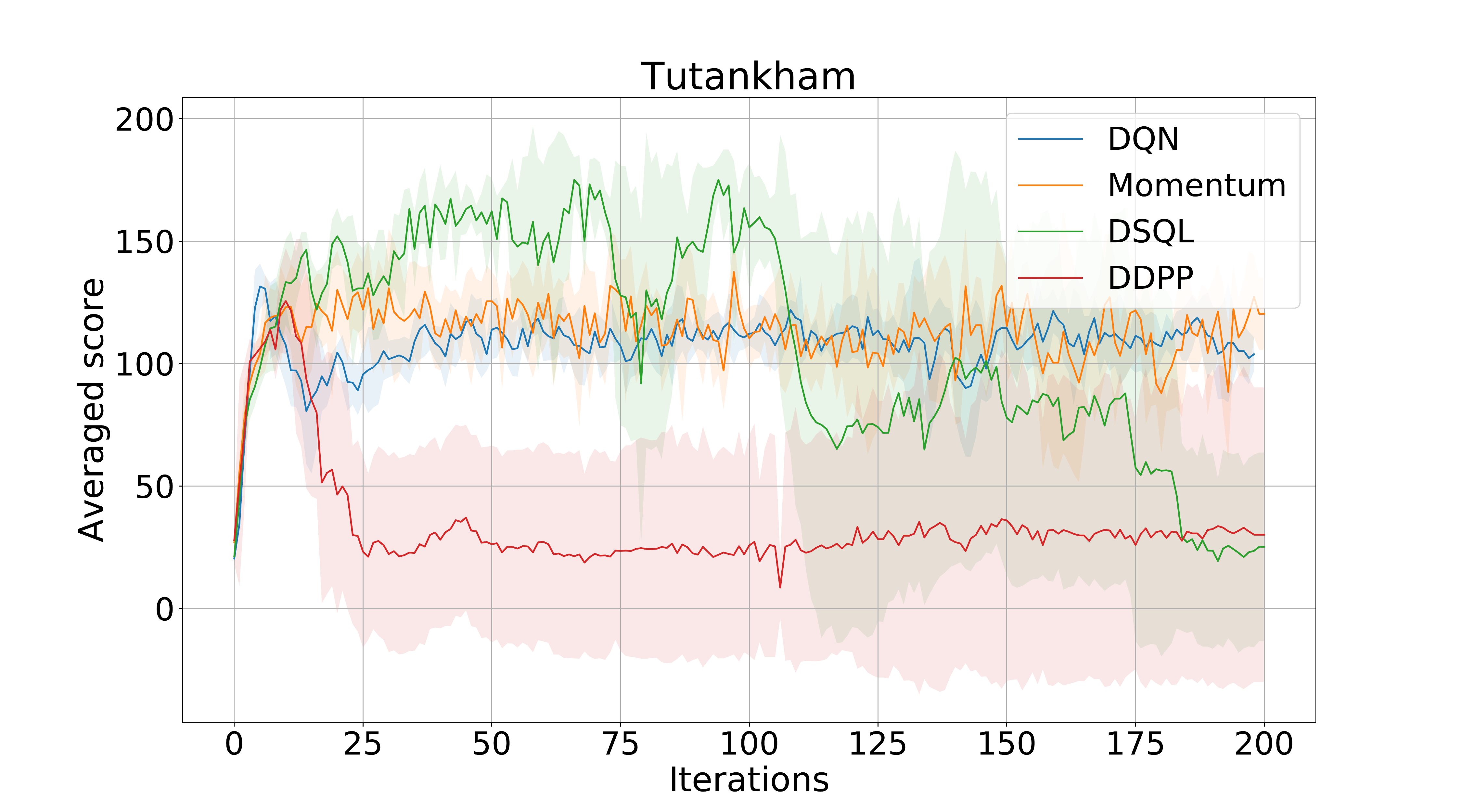}  \\
     SpaceInvaders & Tutankham \\
     \includegraphics[width=0.42\linewidth]{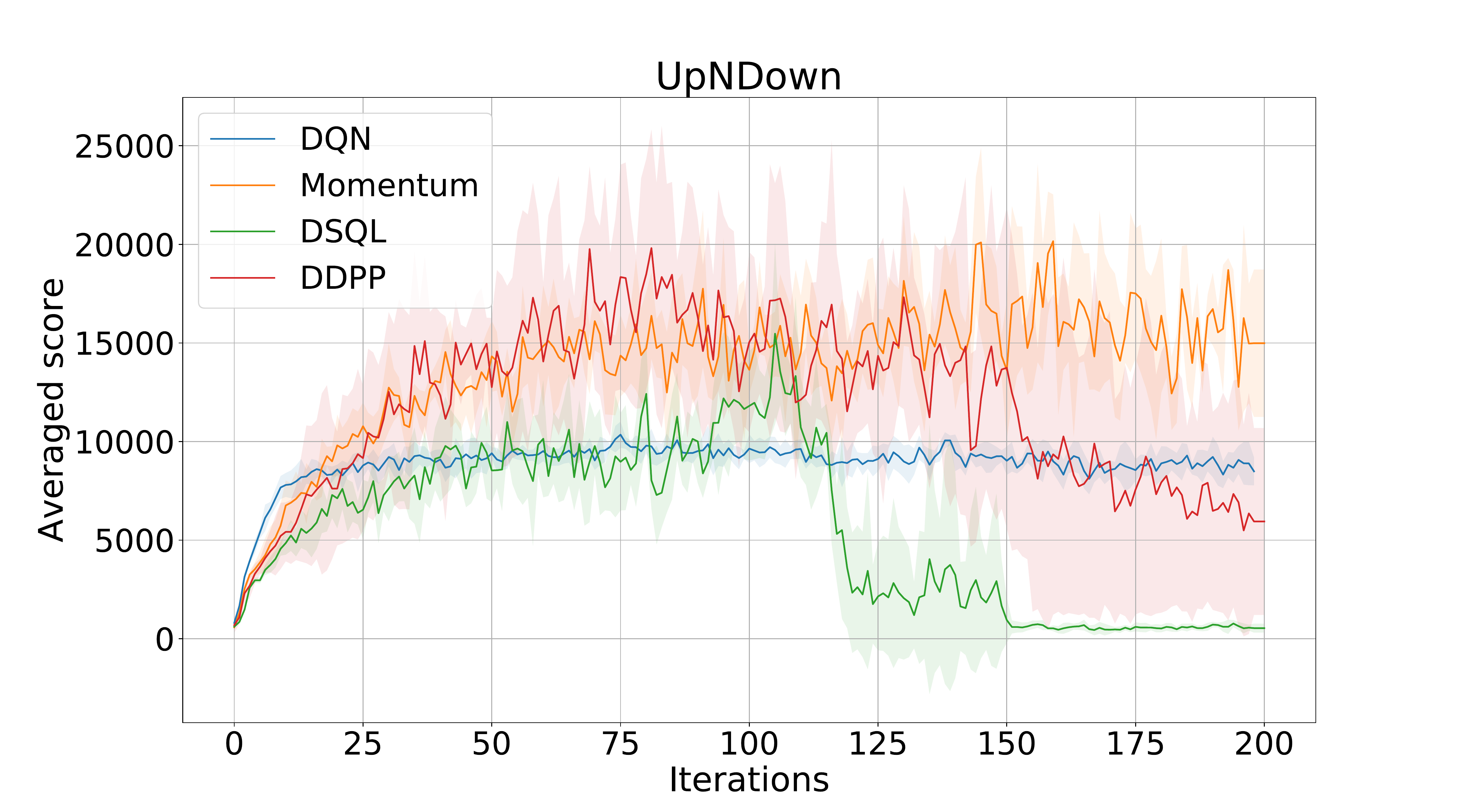} & \includegraphics[width=0.42\linewidth]{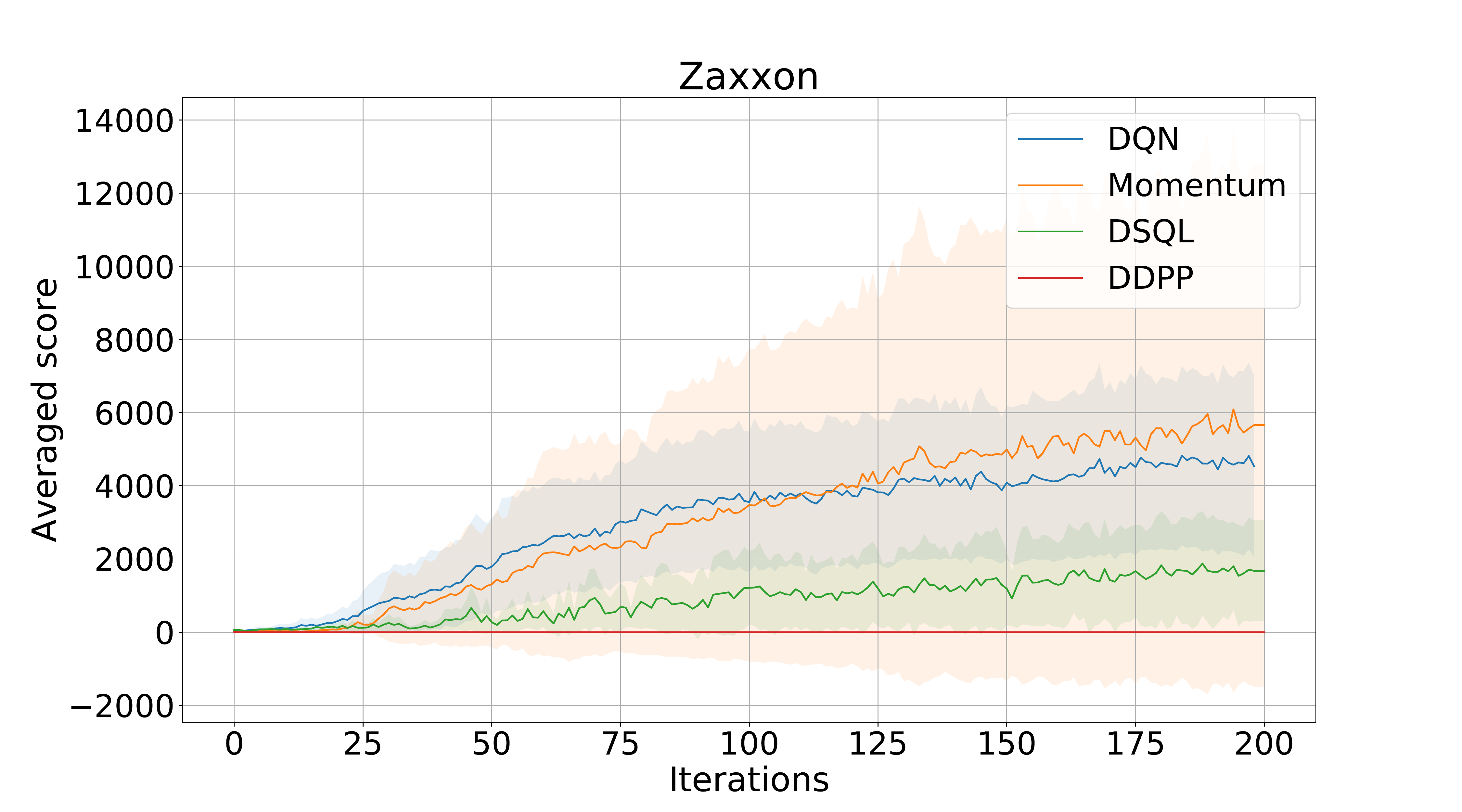}  \\
     UpNDown & Zaxxon \\
\end{tabular}
\caption{All averaged training scores of Momentum-DQN (orange), DSQL (green), DDPP (red), against DQN (blue)  on the subset of Atari games (2/2).\label{fig:full2}}
\end{center}
\end{figure}

\fi

\end{document}